\documentclass{article}

\usepackage[final]{corl_2025} 

\usepackage{graphics,graphicx,float,subcaption,booktabs,xcolor,multirow,array,color,ifthen,tabu,colortbl,dblfloatfix,url,xparse,mathtools,patchcmd,amssymb,xspace,nicefrac,microtype,amsmath,amsthm,amsfonts,bm,ragged2e,tikz,stackengine,etoolbox,xpatch,enumerate,xstring,setspace,tabularx,makecell,changepage,cuted,titlesec,enumitem,wrapfig,tcolorbox,gensymb,algorithm,algpseudocode,makecell,bbm}
\usepackage{titlesec}
\titlespacing*{\section}{0pt}{0.6\baselineskip}{0.5\baselineskip}
\hypersetup{bookmarksopen,bookmarksnumbered,
pdfpagemode=UseOutlines,
colorlinks=true,
linkcolor=red,
anchorcolor=blue,
citecolor=blue,
filecolor=blue,
menucolor=blue,
urlcolor=blue
}
\usepackage[capitalise,noabbrev,nameinlink]{cleveref}
\usepackage[english]{babel}
\usepackage[font=small]{caption}

\newcommand{\link}[1]{\colora{\url{#1}}}

\newcommand{\website}{https://linchangyi1.github.io/LocoTouch}
\newcommand{\ProjectWeb}[0]{\href{\website}{\website}}

\newcommand{\nickname}[0]{LocoTouch\xspace}

\title{LocoTouch: Learning Dynamic Quadrupedal Transport with Tactile Sensing}

\vspace{-0.4cm}
\author{
Changyi Lin\textsuperscript{1}, Yuxin Ray Song\textsuperscript{2}, Boda Huo\textsuperscript{1}, Mingyang Yu\textsuperscript{1}, Yikai Wang\textsuperscript{1}, Shiqi Liu\textsuperscript{1}, \\[0.05cm]
\textbf{Yuxiang Yang\textsuperscript{3}, Wenhao Yu\textsuperscript{3}, Tingnan Zhang\textsuperscript{3}, Jie Tan\textsuperscript{3}, Yiyue Luo\textsuperscript{2}, Ding Zhao\textsuperscript{1}}\\[0.1cm]
\textsuperscript{1}Carnegie Mellon University, \textsuperscript{2}University of Washington, \textsuperscript{3}Google DeepMind\\[0.1cm]
\ProjectWeb
}

\begin{document}
\maketitle

\vspace{-0.8cm}
\begin{figure}[h]
    \centering
    \includegraphics[width= \linewidth]{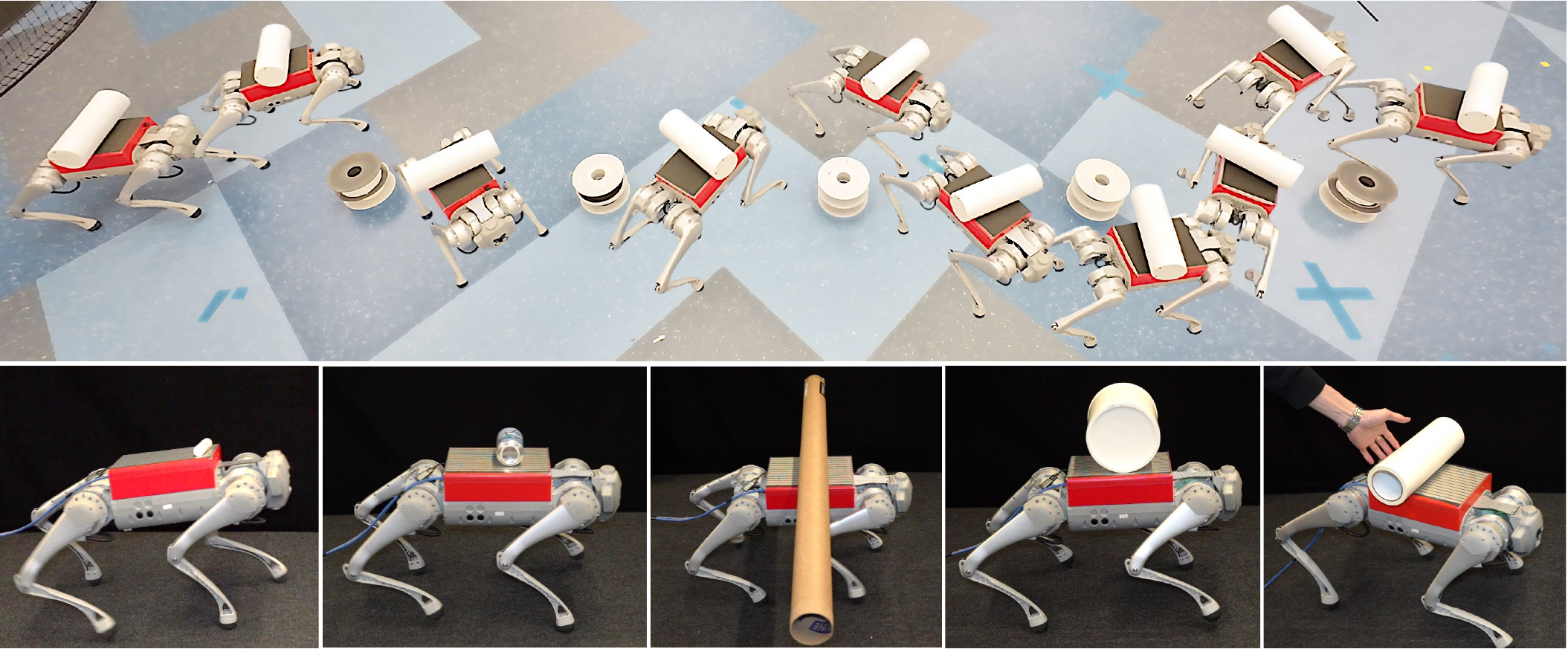}
    \caption{\small We present \nickname, a system for learning tactile-aware quadrupedal policies with zero-shot sim-to-real transfer. Equipped with a custom distributed tactile sensor on the back, \nickname enables quadrupedal robots to robustly transport a wide range of unsecured everyday objects with diverse physical properties.}
    \label{fig:teaser}
\end{figure}

\vspace{-0.3cm}
\begin{abstract}
Quadrupedal robots have demonstrated remarkable agility and robustness in traversing complex terrains.
However, they struggle with dynamic object interactions, where contact must be precisely sensed and controlled.
To bridge this gap, we present LocoTouch, a system that equips quadrupedal robots with tactile sensing to address a particularly challenging task in this category: long-distance transport of unsecured cylindrical objects, which typically requires custom mounting or fastening mechanisms to maintain stability.
For efficient large-area tactile sensing, we design a high-density distributed tactile sensor that covers the entire back of the robot.
To effectively leverage tactile feedback for robot control, we develop a simulation environment with high-fidelity tactile signals, and train tactile-aware transport policies using a two-stage learning pipeline.
Furthermore, we design a novel reward function to promote robust, symmetric, and frequency-adaptive locomotion gaits.
After training in simulation, \nickname transfers zero-shot to the real world, reliably transporting a wide range of unsecured cylindrical objects with diverse sizes, weights, and surface properties.
Moreover, it remains robust over long distances, on uneven terrain, and under severe perturbations.

\end{abstract}

\keywords{Tactile Quadrupedal Policy, Tactile Sim-to-Real, Legged Robots}

\section{Introduction}
\label{sec:intro}

Quadrupedal robots have made remarkable progress in recent years, demonstrating agile and robust locomotion in challenging terrains~\cite{kumar2021rma, agarwal2023legged, margolis2023walk, yang2022fast, margolis2024rapid}.
Building on these advances, researchers have begun exploring object interaction skills for quadrupeds, including button pressing, soccer dribbling, and object pulling~\cite{cheng2023legs, ji2023dribblebot, huang2023creating, fu2023deep, portela2024learning, lin2024locoman, he2025visual}.
Although these behaviors represent important milestones, they remain limited to relatively simple tasks or short-duration interactions.
In contrast, dynamic object interactions that involve sustained contact remain largely unexplored. In this work, we study a particularly challenging task in this category: long-distance transport of unsecured cylindrical objects.

Previous approaches for quadrupedal transport often rely on boxes to contain the object during transport~\cite{valsecchi2023barry, amanzadeh2024predictive, liu2024visual}.
However, such methods limit the variety of object shapes and sizes, cannot adapt the robot action to object movement within the container, and increase the robot's footprint during non-transport tasks, making the system unnecessarily bulky.
For larger objects, secure fastening is particularly challenging and typically requires custom, object-specific mounting mechanisms~\cite{yang2022collaborative, jose2024bilevel, kim2023layered}.
In contrast, transporting objects without containers or mounting mechanisms removes constraints on object geometry and size and eliminates manual fastening procedures.
Despite its benefits, transporting unsecured objects that are dynamically moving entails significant challenges in object state perception, sim-to-real transfer, and ensuring policy robustness and precision.

To address these challenges, we introduce \nickname, a system that equips quadrupedal robots with tactile sensing to enable dexterous transport of unsecured cylindrical objects.
We design a high-density distributed tactile sensor composed of 221 sensing units that cover the robot’s back, enabling efficient perception of dense contact signals.
Inspired by recent successes in visual locomotion policy learning~\cite{agarwal2023legged, cheng2024extreme, zhuang2023robot}, we adapt a two-stage learning pipeline to train a tactile policy in simulation.
During the RL training stage, we design a novel gait to learn stable, symmetric, and frequency-adaptive locomotion gaits for robust object balance and accurate velocity tracking.
To support effective distillation, we introduce a simple yet effective tactile modeling approach that captures the force-conditioned contact spread inherent to the distributed sensor, enabling efficient simulation of high-fidelity tactile signals.
Enabled by these design choices, \nickname transfers zero-shot to the real world (Figure~\ref{fig:teaser}), reliably balancing and transporting a wide range of unsecured, cylindrical everyday objects with diverse weights (0.03–1.45kg), lengths (0.10–1.26m), and diameters (0.03-0.18m). Moreover, it robustly balances objects with slippery surfaces over long distances (tested up to 60m without failure) and under severe external perturbations.

In summary, our main contributions with \nickname are the following:
\begin{itemize}[itemsep=0.2em, topsep=0pt, parsep=0pt]
\vspace{-0.1cm}
    \item A low-cost, high-density, and high-sensitivity distributed tactile sensor that covers the robot's entire back, along with an efficient modeling method for high-fidelity tactile simulation.
    \item An adaptive gait reward that enables the policy to learn stable, symmetric, and frequency-adaptive locomotion gaits without predefined gait patterns and timing references.
    \item The first tactile quadrupedal policy for unsecured object transport that transfers zero-shot to the real world, demonstrating strong generalization, robustness, and agility.
\end{itemize}

\section{Related Works}
\label{sec:related}

\textbf{Quadrupedal Locomotion and Loco-Manipulation.}
Recent advances in quadrupedal locomotion have enabled robots to walk and run on challenging terrains~\cite{kumar2021rma, agarwal2023legged, yang2022fast, margolis2024rapid, cheng2024extreme, zhuang2023robot, rudin2022learning, hoeller2024anymal}.
Beyond locomotion, loco-manipulation has emerged as a promising direction to extend their capabilities to interact with the environment~\cite{cheng2023legs, ji2023dribblebot, huang2023creating, fu2023deep, lin2024locoman}.
However, dynamic interactions involving sustained contact, such as the transport of unsecured objects by quadrupedal robots, remain largely underexplored.

\textbf{Quadrupedal Robots with Tactile Sensing.}
Integrating tactile sensing into quadrupedal robots has attracted increasing interest in the robotics community.
Foot-mounted tactile sensors have been shown to improve locomotion stability and terrain perception~\cite{kaslin2018towards, wu2019tactile, stone2020walking, mudalige2022dogtouch, vangen2023terrain, shi2024foot, song2024tactid}.
To better mimic animal-human interaction, recent work has equipped quadrupedal robots with multiple high-resolution tactile sensors across the torso to collect data for reactive behavior classification~\cite{zhan2023enable, guo2023touch}.
However, these efforts either focus solely on locomotion or limit tactile sensing to passive data collection.
In contrast, our system actively leverages high-density, wide-coverage tactile feedback from the entire back to enable autonomous dexterous transport of unsecured objects.

\textbf{Tactile Simulation and Policy Learning.}
Learning tactile policies from real-world human demonstrations~\cite{yu2023mimictouch, guzey2023dexterity, pattabiraman2024learning, huang20243d, xue2025reactive, zhang2025doglove} has shown promising results, but collecting high-quality tactile demonstrations is often difficult and expensive.
A more efficient and scalable way is to learn train in simulation.
For point-based tactile sensors, characterized by discrete sensing elements without signal interference, each sensor is typically simulated as an individual rigid body, and manipulation policies are learned based on binary contact signals~\cite{yin2023rotating, yuan2024robot, xue2024arraybot, yin2024learning}.
For distributed tactile sensors, advanced simulation methods have been developed to model material softness and force spread accurately~\cite{narang2020interpreting, narang2021sim, kappassov2020simulation, kasolowsky2024fine, leins2025hydroelastictouch}, but these approaches are often computationally expensive for the learning of RL policies, which requires massively parallel simulation environments.
To improve efficiency, compliant contact models have been used to simulate coarse binary signals~\cite{ding2021sim, yang2023tacgnn}, which do not capture the contact spread of real signals.
In contrast, our expanded collision modeling efficiently simulates realistic contact spread.
Furthermore, the tactile policies mentioned above focus mainly on robotic arms or dexterous hands.
In contrast, our task requires a quadrupedal robot, a highly dynamic system, to perform simultaneous locomotion and object interaction using dense tactile input (221 taxels), posing unique challenges for both policy learning and sim-to-real transfer.


\section{Overview of \nickname}
\begin{figure}[h]
  \centering
  \includegraphics[width= \linewidth]{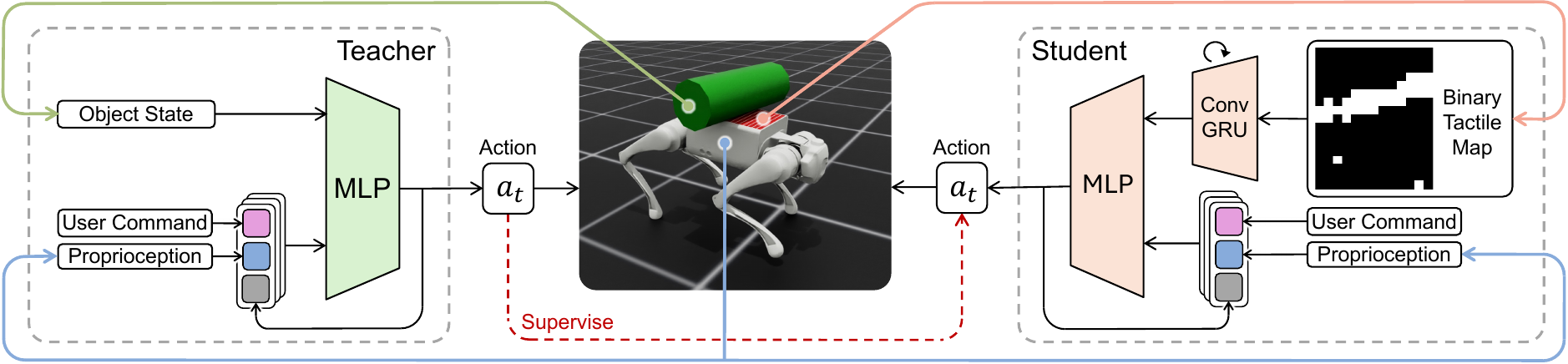}
  \caption{\small Training pipeline of \nickname. The teacher policy is trained via reinforcement learning using privileged object state information. The student policy replaces object state with tactile observations encoded by a Conv-GRU network and is distilled from the teacher using Behavior Cloning and DAgger. Both policies share proprioception and command inputs, and output joint position targets for the robot.
  }
  \label{fig:pipeline}
\end{figure}

Our goal is to learn a tactile policy for quadrupedal robots to balance and transport unsecured objects on the back in the real world. To enable this, we design a high-density distributed tactile sensor, simulate its taxels efficiently and accurately, and process the tactile signals to enable zero-shot sim-to-real transfer (Section~\ref{sec:tactile_sensing}). For learning an effective policy with tactile input, we adapt a two-stage learning method (Figure~\ref{fig:pipeline}) by firstly training the teacher policy with object state and proprioception sequences as inputs
using Proximal Policy Optimization (PPO)~\cite{schulman2017proximal}, and utilizing DAgger~\cite{ross2011reduction} to distill it to the student policy that observes the tactile signals for real world deployment (Section~\ref{sec:policy_learning}).

\section{Distributed Tactile Sensing and Simulation}
\begin{figure}
  \centering
  \includegraphics[width=\linewidth]{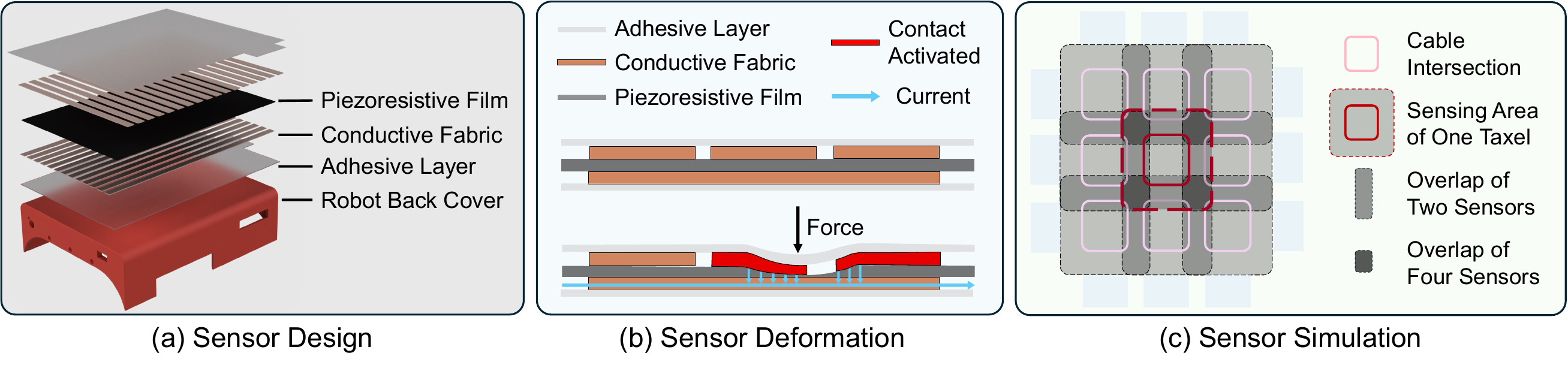}
  \caption{\small Distributed tactile sensing for \nickname. (a) Our custom, high-density tactile sensor covers the entire back of the robot. (b) Force propagation through soft materials activates adjacent taxels. (c) The collision model for each taxel is expanded in simulation to simulate the spatially coupled contacts in real sensors.}
  \label{fig:sensor}
\end{figure}

\paragraph{Sensor Design}
As shown in Figure~\ref{fig:sensor}~(a), we design a high-density, large-area tactile sensor that fully covers the back cover, which is mounted on the torso of a quadrupedal robot.
The tactile sensor consists of a piezoresistive film, sandwiched between orthogonally aligned conductive electrodes made of conductive fabric. We equip the sensing array with additional layers of adhesive at the top and bottom for fastening and protection. Each tactile unit, or \textit{taxel}, is located at the intersection of conductive electrodes.
When external contact forces deform the piezoresistive film, the resulting change in resistance is detected via the intersecting electrodes.
We can measure the resistance at each individual taxel using an electrical-grounding-based signal isolation circuit ~\cite{sundaram2019learning}.
The taxels are arranged in a distributed grid with an overall coverage of $\text{250} \times \text{180}~\mathrm{mm}$, divided into 17 rows on the top layer and 13 columns on the bottom layer.
The intersections of each row and column create an evenly distributed array of 221 taxels, where each intersection block covers an effective sensing area of $\text{14.3}\times\text{12.8}~\mathrm{mm}$.
To acquire signals efficiently, we sequentially process the tactile signals one row at a time, where we apply a constant current from the top layer, and measure the voltage drop at the bottom layer.
For reliable tactile sensing with wide signal coverage, we minimize the gap between adjacent taxels, and design each tactile unit to be $\text{11.3}\times\text{10.5}~\mathrm{mm}$ in size.

\label{sec:tactile_sensing}

\paragraph{Simulation of Coupled Contacts}
While the matrix-based tactile sensor design improves scalability and wiring efficiency, it inevitably introduces signal coupling between adjacent taxels. As illustrated in Figure~\ref{fig:sensor}~(b), when a force is applied to a taxel, the mechanical deformation caused by the contact force tends to propagate through the soft sensing material, resulting in contact signals in adjacent taxels as well.
Moreover, this contact spread is highly sensitive to the precise position of contact.
For example, compared to a force applied at the center, a force applied close to the edge of a taxel is much more likely to trigger force response in nearby cells.
Therefore, standard, taxel-based approaches like convolutional filters cannot accurately capture the nature of such contact coupling.

To efficiently reproduce this coupling of contact signals in simulation, we create the simulated taxels with a calibrated enlarged collision area of $\text{18.3}\times\text{17.5}~\mathrm{mm}$ for each unit, so that contact forces applied closer to the taxel edge trigger sensor readings in multiple nearby taxels, as demonstrated in Figure~\ref{fig:sensor}~(c).
Without explicitly querying and processing the exact position of each contact point, this approach enables fast and scalable simulation of realistic tactile signal patterns, capturing both localized and distributed contact characteristics. Please refer to Appendix~\ref{sec:app_tactile} for calibration details of the expanded model.

\paragraph{Binary Contact Signals}
While ground-truth contact forces is directly available in simulation, real-world tactile sensors estimate these forces from voltage readings, where the voltage-force relationship is noisy and highly non-linear.
Even with careful calibration, the exact values of the contact forces may still differ significantly from simulation, creating a significant sim-to-real gap.
To address this, we convert the contact signals to binary 0-1 readings~\cite{yin2023rotating, xue2024arraybot}, where the force threshold is set to be a fixed number in the simulation ( $f_{\text{sim}}=0.05N$), and a calibrated voltage change value in the real sensor.
To further narrow the sim-to-real gap, we randomly flip $0.5\%$ of contact and non-contact signals respectively in simulation.

\section{Tactile-Aware Quadrupedal Policy Learning}
\label{sec:policy_learning}

\subsection{Overview of Teacher-Student Approach}
Similar to previous works \cite{kumar2021rma, agarwal2023legged, zhuang2023robot}, we adopt a teacher-student approach to train the tactile policy.
In the first step, we train the teacher policy using reinforcement learning (RL), where the teacher policy observes object states (Section~\ref{sec:teacher}).
We then distill the teacher policy into a student policy using DAgger, where the student policy imitates the behavior of the teacher policy with simulated tactile inputs (Section~\ref{sec:student}).

\subsection{Teacher Policy Training}
\label{sec:teacher}
\paragraph{State and Action Space}
We model the environment of the teacher policy as a Markov Decision Process (MDP), where the observation space $o_t\in\mathbb{R}^{58}$ consists of the pose and velocity of the object $s_t^{\text{obj}}\in\mathbb{R}^{13}$ (position, quaternion, linear velocity and angular velocity) in the robot frame, robot proprioception $s_t^{\text{proprio}}\in\mathbb{R}^{30}$ (gravity vector, base angular velocity, joint positions and velocities), velocity command $v_t^{\text{cmd}}\in\mathbb{R}^{3}$, and the last action $a_{t-1}\in\mathbb{R}^{12}$. 
The teacher policy takes a $\text{6}$-step history of observations $o_{(t-H) \dots t}\in\mathbb{R}^{340}$ as input and outputs the target joint positions $a_t \in \mathbb{R}^{12}$, which is tracked by a following PD controller.

\paragraph{Environment for Object Transport}
In our environment, the robot transports a diverse set of cylindrical objects with randomized physical properties. Specifically, we randomize each object's radius \([0.03\,\mathrm{m},\, 0.07\,\mathrm{m}]\), length \([0.1\,\mathrm{m},\, 0.4\,\mathrm{m}]\), mass \([0.5\,\mathrm{kg},\, 2.5\,\mathrm{kg}]\), and contact friction coefficient \([0.3,\, 1.0]\). The initial object pose is also randomized, with translations along \(x\) and \(y\) sampled from \([-0.05\,\mathrm{m},\, 0.05\,\mathrm{m}]\) and \([-0.04\,\mathrm{m},\, 0.04\,\mathrm{m}]\), respectively, and yaw from \([-30^\circ,\, 30^\circ]\).
To ensure a stable starting phase and prevent premature policy reactions, we set all velocity commands to zero during the first 50 steps of each episode and mask the object state observations with Gaussian noise until the object makes its first contact with the robot. For training efficiency, we terminate episodes early if the object falls beneath the robot's torso or if the torso or hips contact the ground.
To improve learning robustness, we apply random external pushes to both the object and the robot during training. We also adopt a velocity command curriculum that adapts based on velocity tracking performances.

\paragraph{Reward Design}
We design the reward to consist of 5 components:
\vspace{-0.15cm}
\begin{align*}
    r = r_{\text{alive}} + r_{\text{task}} + r_{\text{drag}} + r_{\text{gait}} + r_{\text{reg}}
\end{align*}
\vspace{-0.55cm}
\\
$r_{\text{alive}}$ is a constant that ensures the overall reward remains positive.
$r_{\text{task}}$ encourages the robot to track the velocity command and balance the object.
$r_{\text{reg}}$ is a set of regulatory terms for stable and smooth locomotion \cite{margolis2023walk}.
To mitigate the sim-to-real gap caused by foot deformation and enable stable deployment in the real world, we define $r_{\text{drag}}$ as a sum of four binary penalty terms, one for each leg.
Each penalty activates when a foot is near the ground while exhibiting high lateral velocity (details in Appendix~\ref{sec:app_reward}).
Finally, $r_{\text{gait}}$ is an adaptive term that promotes symmetric and stable gait patterns. 
We find this term to be essential for learning a robust and reliable transport policy, and we discuss it in detail in Section~\ref{sec:gait}.

\subsection{Adaptive Gait Reward}
\label{sec:gait}

While several locomotion gaits have been implemented in quadrupedal robots \cite{margolis2023walk, yang2022fast}, the \emph{trotting} gait remains the preferred choice for stable object transportation.
To promote trotting behavior, prior work~\cite{margolis2023walk, liu2024visual} introduces contact references into the reward function, typically based on fixed stepping frequencies.
However, imposing fixed contact sequences and explicit timing constraints prevents the policy from learning adaptive locomotion behaviors, which is particularly important in our case of unsecured object transport.
In this work, we introduce an \emph{adaptive} gait reward that removes the need for any externally provided timing reference, allowing the policy to learn flexible and dynamic gait timing in a \emph{self-supervised} manner.

Let $c_i=\{0, 1\}$ denote the contact state of the $i$th leg, where $i={1,\dots,4}$ corresponds the front-right (FR), front-left (FL), rear-right (RR), and rear-left (RL) legs, respectively.
The 4 legs form 6 pairs, which we group into diagonal pairs $P^{\text{diag}}=\{(FR, RL), (FL, RR)\}$ and lateral pairs $P^{\text{lat}}=\{(FR, FL), (FL, RL), (RL, RR), (RR, FR)\}$~\cite{mittal2023orbit}.
Given the alternating contact of diagonal pairs in the trotting gait, we define the adaptive gait reward as:
\begin{align}
    r_{\text{gait}} = \frac{1}{2}\sum_{(i,j)\in P^{\text{diag}}}\gamma_{\text{sym}}\mathbf{1}_{\{c_i=c_j\}}
    + \frac{1}{4}\sum_{(i,j)\in P^{\text{lat}}}\mathbf{1}_{\{c_i\neq c_j\}}
\end{align}
\vspace{-0.33cm}

where the first term encourages synchronized contact states between diagonal legs, and the second term promotes alternating contact states between lateral legs. To further guide the learning of symmetric air times and encourage a relatively low gait frequency, we define an adaptive coefficient $\gamma_{\text{sym}}$ for the diagonal term. We briefly discuss the construction of $\gamma_{\text{sim}}$ here, and include the detailed formulation in Appendix~\ref{sec:app_reward}. We define $\gamma_{\text{sim}}$ as:
\begin{align}
\gamma_{\text{sym}}=\begin{cases}
    1 & c_i=c_j=1\\ \alpha_{\text{task}}f_{\text{sym}}\left(t_{\text{curr}}|t_{\text{prev}}, t'_{\text{prev}}\right) & c_i=c_j=0
\end{cases}
\label{equ:symmetry_weight}
\end{align}
\vspace{-0.33cm}

In short, $\gamma_{\text{sym}}$ takes a constant value when the diagonal pair is in contact and an evaluated symmetricity score when both legs are in swing. More specifically, for a diagonal pair in swing ($c_i=c_j=0$), we define $\gamma_{\text{sym}}$ as the product of a normalized task performance score $\alpha_{\text{task}}\in[0,1]$ and a symmetricity score $f_{\text{sym}}\left(t_{\text{curr}}|t_{\text{prev}}, t'_{\text{prev}}\right)\in[-1,1]$.
The task performance score $\alpha_{\text{task}}$ evaluates the robot’s performance in velocity tracking and object balancing.
\begin{wrapfigure}{r}{0.4\textwidth}
    \vspace{-0.5cm}
    \centering
    \includegraphics[width=\linewidth]{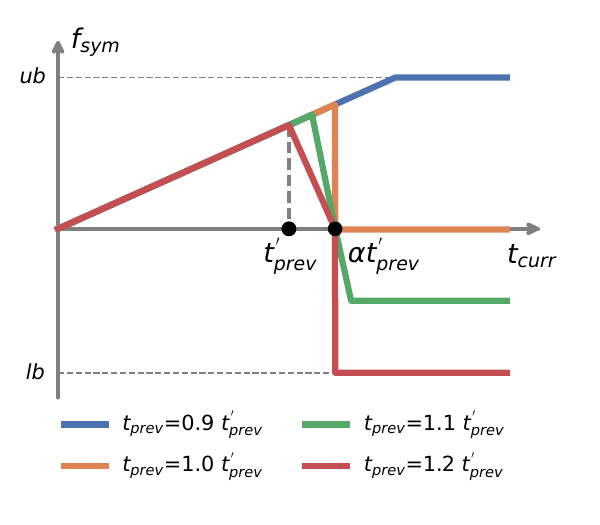}
    \vspace{-0.7cm}
    \caption{\small Symmetricity score $f_{\text{sym}}$ for varying previous air times $t_{\text{prev}}$. When $t_{\text{prev}} < t'_{\text{prev}}$, $f_{\text{sym}}$ increases linearly up to an upper bound, encouraging longer air times and a more stable gait. When $t_{\text{prev}} > t'_{\text{prev}}$, $f_{\text{sym}}$ penalizes excessively long swing durations, promoting more symmetric gaits.
    }
    \vspace{-0.7cm}
    \label{fig:gait_reward}
\end{wrapfigure}
The symmetricity score is computed based on $t_{\text{curr}}, t_{\text{prev}}$, $t'_{\text{prev}}$, which denote the air time of the current ($t$) or the alternative diagonal pair ($t'$) of feet, in the current or the previous gait cycle.
We plot $f_{\text{sym}}$ for different values of $t_{\text{prev}}$ and $t'_{\text{prev}}$ in Figure~\ref{fig:gait_reward}. Intuitively, if the current pair encountered shorter previous air time $(t_{\text{prev}}<t'_{\text{prev}})$ as the blue line show, $f_{\text{sym}}$ encourages it to have longer air time with the score symmetricity up to an upper bound $ub$.
If the current pair experienced longer previous air time $(t_{\text{prev}}\geq t'_{\text{prev}})$, $f_{\text{sym}}$ encourages current air time up to $\alpha\,t'_{\text{prev}}$, and penalizes exceedingly long air times, where $\alpha$ is $1.2$ in our implementation. Furthermore, the penalty is more serious if their previous air time difference is larger, as demonstrated with the other three lines.
Since this symmetricity score is computed individually for both diagonal pairs, it encourages the foot pairs to have long swing times while respecting the previous swing time of the alternative pair.

\subsection{Student Policy Distillation}
\label{sec:student}
After training the teacher policy that relies on object state obtained from simulation, we aim to train a student policy based on onboard tactile sensing for deployment. The only input difference between the teacher and student policies is that the student policy observes tactile map while the teacher policy observes object state sequence. As shown in Figure~\ref{fig:pipeline}, following a similar monolithic architecture as in~\cite{agarwal2023legged}, we use a convnet-GRU encoder to encode the tactile map and feed the tactile embedding, along with other observations, into the MLP-based policy backbone. We adopt a two-stage distillation process~\cite{yang2024agile}, first collecting a dataset of 400k state-action pairs from the teacher to pre-train the student policy via Behavior Cloning (BC). We then refine the student policy using DAgger~\cite{ross2011reduction}, training it over 7 iterations, with each iteration collecting 200k additional state-action pairs.  To better match real-world conditions, we also measure the tactile signal latency ($0.02\,\text{--}\,0.04\text{s}$) and simulate this delay during training.

\section{Experiments}
\label{sec:result}
We design experiments to answer the following questions: 1) Is \nickname able to achieve robust object transport in the real-world in face of changing commands and object sliding (Sec. \ref{sec:system_performance})?  2) How important is the inclusion of tactile sensing in these tasks (Sec. \ref{sec:quant})? 3) How does the proposed taxel modeling method affect the simulation fidelity (Sec. \ref{sec:simulation_result})? 4)How does incorporating the symmetricity function alter the gait behavior (Sec. \ref{sec:gait_result})?

\subsection{Dexterous Quadrupedal Transport with Tactile Sensing}
\label{sec:system_performance}

\textbf{\nickname enables general, robust object transport.}
We equip the Unitree Go1 robot’s back with our custom distributed tactile sensor, as shown in Figure~\ref{fig:teaser}.
The tactile policy enables the robot to dynamically balance a 1.45kg cylinder and successfully transport it through dense obstacles under frequent sharp turns, demonstrating its ability to track velocity commands and maintain object stability in complex real-world scenarios.
Furthermore, the robot can reliably balance and transport a diverse set of everyday objects with significantly varying, out-of-distribution geometries -- ranging in diameter from 0.03 to 0.18m, length from 0.10 to 1.26m, and mass from 0.03 to 1.45kg, including items with extremely slippery surfaces, such as a drink can.
It also maintains stable transport even under external perturbations and disturbances, as demonstrated in the accompanying \href{\website}{video}.

\textbf{\nickname exhibits agile and adaptive behaviors.}
To evaluate \nickname's agility and adaptability during object transport, we apply time-varying forward velocity commands and record both robot and object states using a motion-capture system and onboard foot-contact sensors.
\begin{wrapfigure}{r}{0.55\textwidth}
    \vspace{-0.4cm}
    \includegraphics[width=\linewidth]{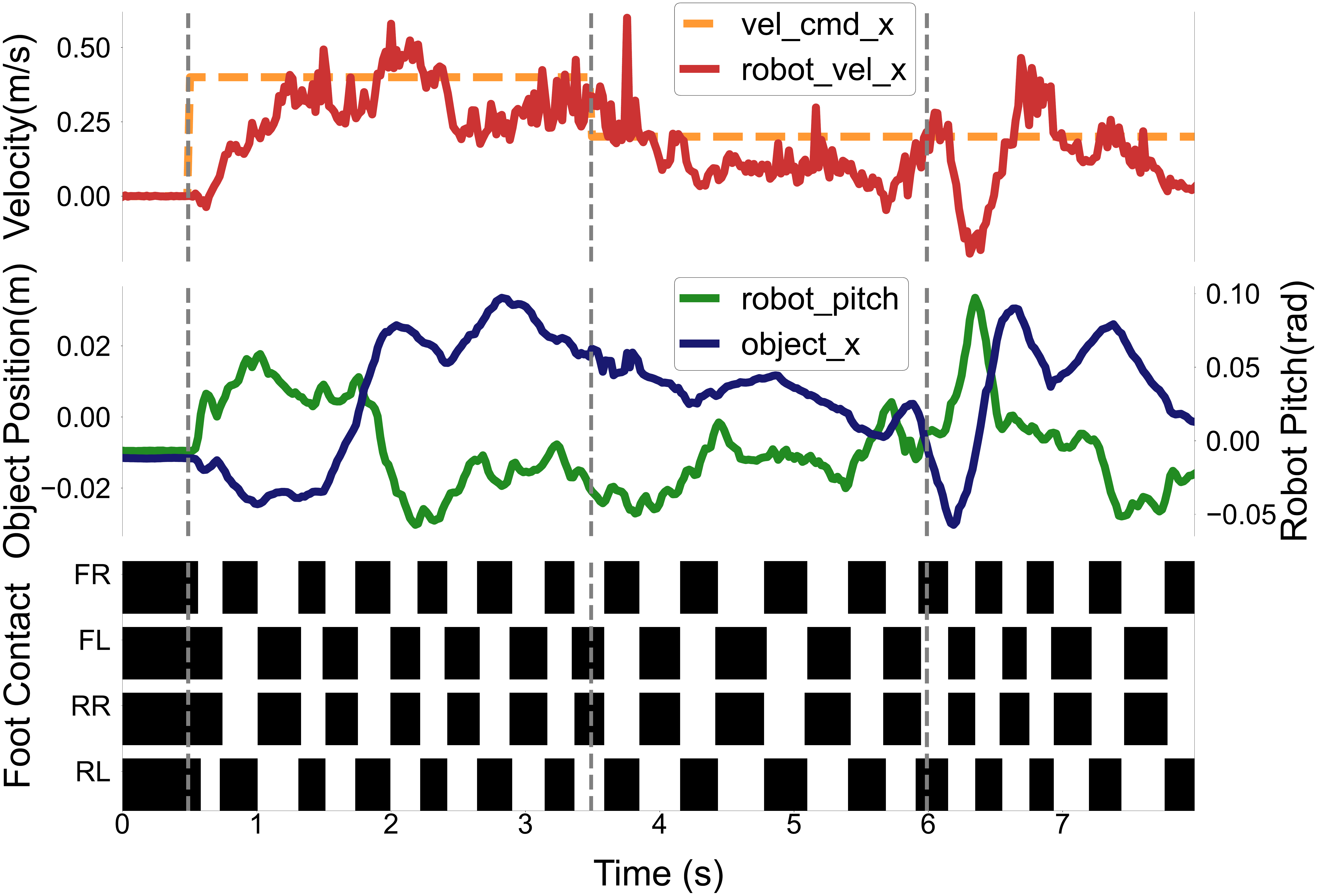}
    \vspace{-0.6cm}
    \caption{
    Transport object under time-varying forward velocity commands. \textbf{Top:} commanded velocity and measured robot velocity. \textbf{Middle:} robot pitch and object forward displacement in the robot frame. \textbf{Bottom:} binary foot-contact patterns, where black segments indicate contact. }
    \label{fig:tracking_combined}
    \vspace{-0.5cm}
\end{wrapfigure}
As shown in Figure~\ref{fig:tracking_combined}, the robot rapidly responds to track the sharp velocity command changes (e.g., from 0 to 0.4 m/s at 0.5 s and from 0.4 to 0.2 m/s at 3.5 s), while maintaining object balance. Notably, we observe an inverse coupling between the robot’s pitch angle and the object’s forward displacement in the robot frame, indicating active compensation for object motion.
Furthermore, the foot-contact pattern reveals adaptive stepping: the robot slows its stepping frequency when the velocity command drops at 3.5 s, then speeds up again at 6.0 s in response to the object sliding backward, even though the commanded velocity remains unchanged. These behaviors highlight the policy’s ability to adaptively adjust the gait frequency based on both commanded velocity and object state.

\subsection{Quantitative Evaluation of \nickname}
\label{sec:quant}
We evaluate the performance of \nickname in simulation on two task families: pure balancing (B), which uses zero-velocity commands, and object transport (T), which uses non-zero velocity commands. For each setting, we assess whether the policy can successfully complete the task for 5s and 10s, and compute the success rate over 8,000 trajectories. All evaluations are conducted in the same environment used during training, with identical terrain, robot, and object configurations. Table~\ref{tab:success_rate} reports the resulting success rates for five policy variants:
\begin{wraptable}{r}{0.55\textwidth}
    \vspace{-0.35cm}
    \resizebox{\linewidth}{!}{
    \begin{tabular}{lcccc}
        \toprule
        \multirow{2}{*}{Policy Variant} & \multicolumn{4}{c}{Success Rate} \\
        \cmidrule{2-5} & B-5s & B-10s & T-5s & T-10s \\
        \midrule
        Teacher& 99.4\% & 98.8\% & 99.8\% & 99.8\% \\
        Teacher w/o Object State & 0.0\% & 0.0\% & 0.0\% & 0.0\% \\
        Student(\emph{ours}) & 96.2\% & 95.8\% & 96.7\% & 96.4\% \\
        Student w/o Tactile Sensing & 0.0\% & 0.0\% & 0.0\% & 0.0\% \\
         Locomotion Only & 0.77\% & 0.42\% & 0.20\% & 0.03\% \\
        \bottomrule
    \end{tabular}}
    \vspace{-0.15cm}
    \caption{\small Success rates of different policies on balancing (B) and transport (T) tasks, measured over 5-second and 10-second durations.}
    \vspace{-0.5cm}
    \label{tab:success_rate}
\end{wraptable}
(a) the teacher policy in our training pipeline;  
(b) a teacher variant that omits object-state inputs;  
(c) the resulting student policy;
(d) a student variant without tactile inputs; and
(e) a locomotion-only policy trained without object interaction.

The teacher policy achieves near-perfect success, and our tactile student policy similarly maintains high performance across all tasks. By contrast, the locomotion-only baseline and the teacher variant without object-state inputs perform poorly, demonstrating that rich object-interaction feedback is indispensable. Removing tactile input from the student also leads to complete failure, underscoring the critical role of tactile sensing in enabling object balancing and transport. We also perform quantitative evaluations of \nickname in real-world scenarios; details are provided in Appendix~\ref{sec:app_exp_diverse_obj}.

\subsection{High-Fidelity Tactile Simulation with Expanded Collision Model}
\label{sec:simulation_result}
To evaluate the fidelity of our tactile simulation, we collect tactile signals from the real sensor while tracking a cylindrical object’s pose relative to the robot frame via motion capture.
We then replay these poses in simulation to generate tactile signals using three taxel simulation models:
\begin{wrapfigure}{r}{0.5\textwidth}
    \centering
    \includegraphics[width=\linewidth]{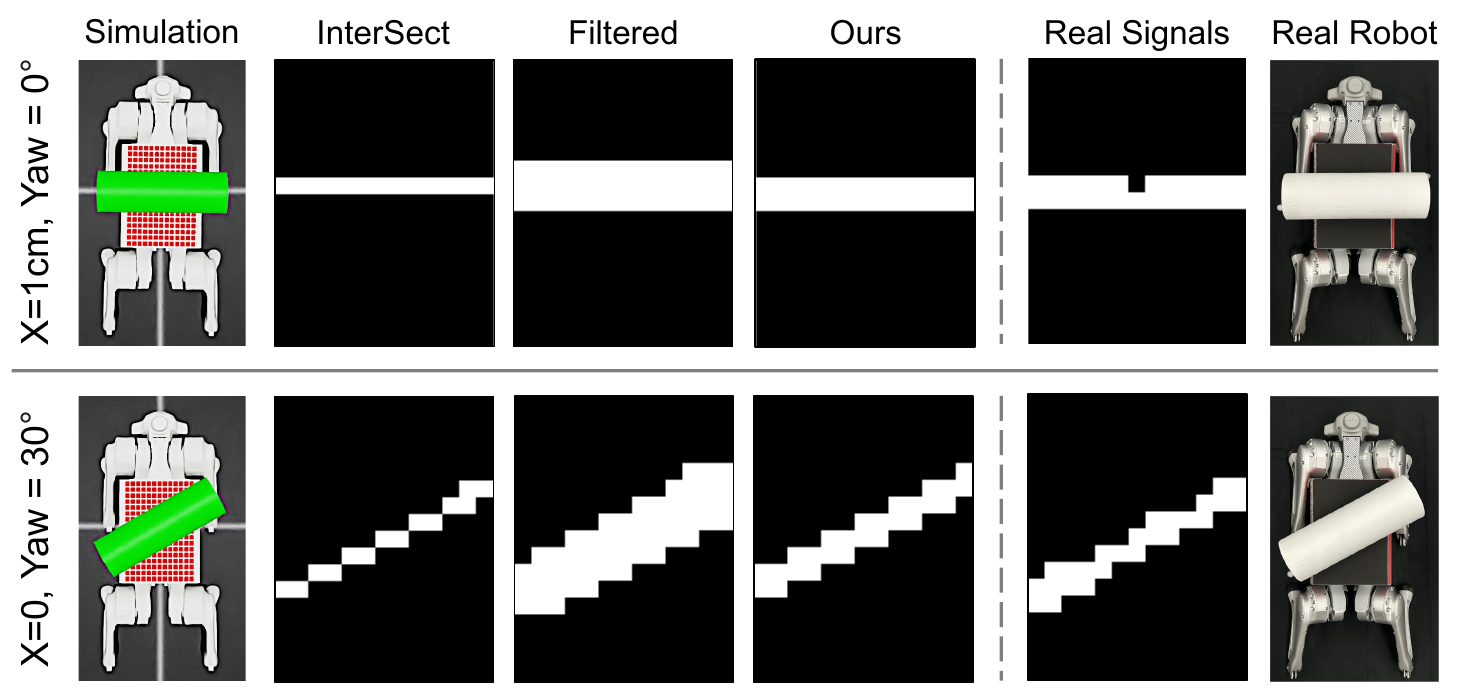}
    \vspace{-0.45cm}
    \caption{\small Comparison of tactile maps generated by different simulation methods with real sensor signals. Object poses are ($x = 0.01\,\mathrm{m},\ y = 0,\ \text{yaw} = 0^\circ$) and ($x = 0,\ y = 0,\ \text{yaw} = 30^\circ$).}
    \vspace{-0.3cm}
    \label{fig:tactile_sim_to_real}
\end{wrapfigure}
(a) \textbf{InterSect}: an intersection-only model that detects contact only at cable intersections; (b) \textbf{Filtered}: a Gaussian-smoothed model that applies a Gaussian filter (kernel size 3) to the output of InterSect, and (c) \textbf{Ours}: the proposed expanded collision model that enlarges each taxel's effective contact area. Figure~\ref{fig:tactile_sim_to_real} compares the simulated and real signals for two representative object placements. As shown in the figure, the expanded contact model reproduces the sensor contact patterns with the most faithfulness in both shape and spatial distribution. In contrast, InterSect underestimates the contact spread, while the Filtered model excessively blurs it.

\subsection{Symmetricity Function Improves Gait Symmetry and Task Performance}
\label{sec:gait_result}
\begin{wrapfigure}{r}{0.4\textwidth}
    \vspace{-0.55cm}
    \centering
    \includegraphics[width=\linewidth]{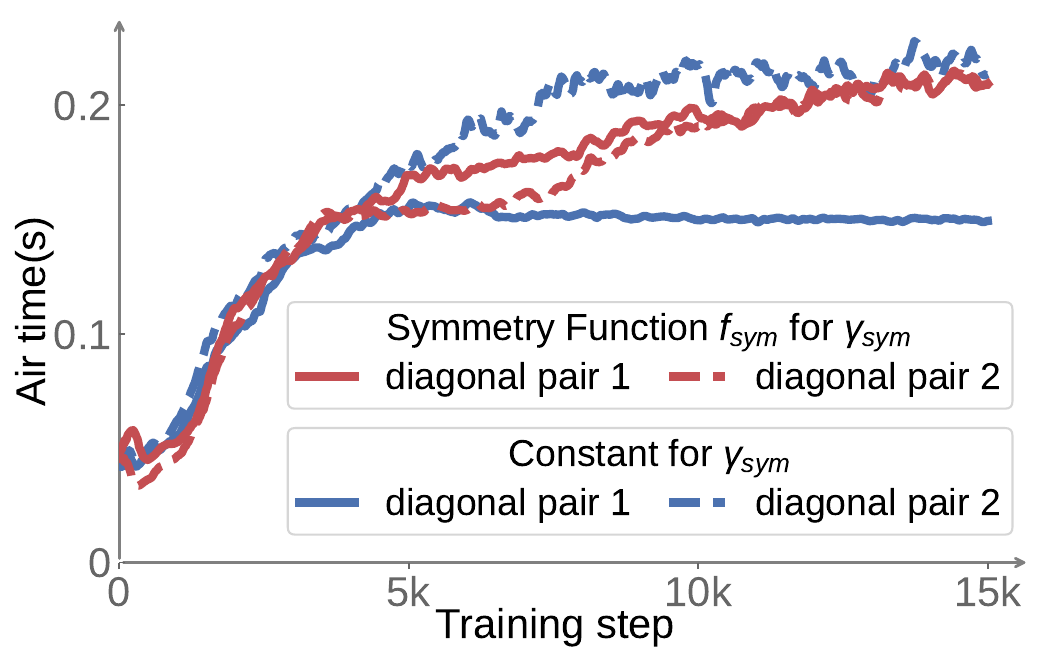}
    \vspace{-0.6cm}
    \caption{\small Comparison of stepping air time for diagonal foot pairs over training. Our method promotes low-frequency symmetric gaits.}
    \label{fig:stepping_frequency}
    \vspace{-0.5cm}
\end{wrapfigure}
As described in Section~\ref{sec:system_performance}, the robot rapidly adapts its stepping frequency in response to command changes or object sliding while maintaining gait symmetry.
We further compare two approaches for setting $\gamma_{\text{sym}}$ in the gait reward $r_\text{gait}$ during teacher policy training: a constant value ($\gamma_{\text{sym}}=1$) as the baseline, and our symmetricity function (Equation~\ref{equ:symmetry_weight}).
To evaluate gait symmetry, we track the stepping air time of the two diagonal foot pairs.
As shown in Figure~\ref{fig:stepping_frequency}, the baseline policy exhibits a significant discrepancy in air time between the diagonal pairs, indicating poor symmetry.
In contrast, our symmetricity function yields well-aligned air times across both pairs and encourages a lower overall stepping frequency.

\begin{figure*}[h]
    \vspace{-0.3cm}
    \centering
    \includegraphics[width=0.97\linewidth]{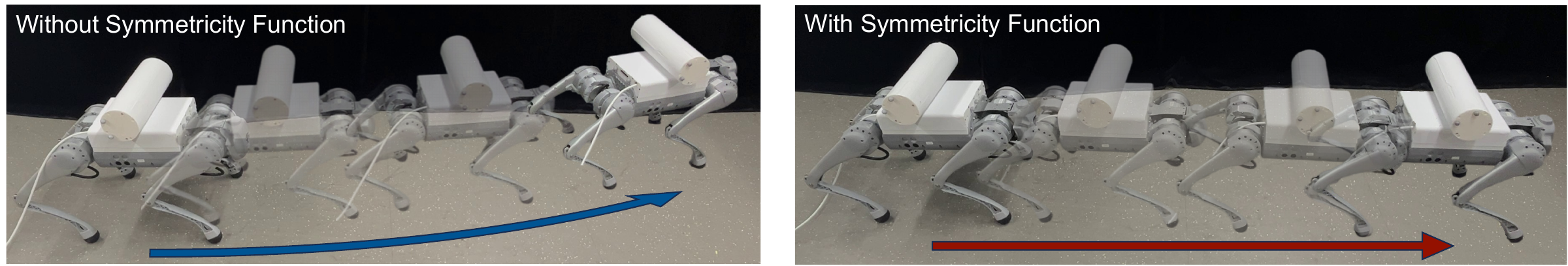}
    \vspace{-0.05cm}
    \caption{\small Baseline teacher policy exhibits lateral drifting due to asymmetric deformation of the foot pairs; while policy trained with our symmetricity function tracks the forward velocity command accurately.}
    \label{fig:symmetricity_vel_tracking}
\end{figure*}

Notably, although both teacher policies perform similarly in simulation, their real-world behaviors differ significantly.
As shown in Figure~\ref{fig:symmetricity_vel_tracking}, we deploy both policies on real robots with a forward velocity command of $0.3\mathrm{m/s}$.
The baseline policy exhibits asymmetric gaits, which not only appears unnatural but also degrades tracking performance due to uneven foot deformation across the diagonal pairs. In contrast, our policy maintains symmetric gaits and accurately tracks the commanded velocity.

\vspace{-0.3cm}
\section{Conclusion}
\vspace{-0.1cm}
We presented \nickname, a quadrupedal system that enables dynamic and dexterous transport of unsecured objects through distributed tactile sensing and learning-based control.
Our approach features three key components: a high-density tactile sensor, an efficient tactile simulation model that generates high-fidelity signals, and an adaptive gait reward that facilitates symmetric and adaptive gaits.
Together, they enable \nickname to transfer zero-shot to the real world, reliably balancing and transporting diverse unsecured everyday objects with strong generalization, robustness, and agility.

In the future, we plan to extend \nickname with multi-modal sensing such as visual-tactile sensing to support more versatile quadrupedal skills. In addition, expanding tactile coverage to the full body opens up exciting opportunities for omnidirectional interaction and whole-body manipulation.
\clearpage

\section{Limitations}
\vspace{-0.2cm}
Our system encounters difficulties when balancing cylindrical objects whose axes are aligned with the robot's forward torso direction. This limitation comes from two main factors: (1) the training setup only randomizes the object’s yaw angle within \([-30^\circ,\,30^\circ]\) owning to the assumption of placing the object near \(0^\circ\), which makes longitudinal placements out of distribution for the learned policy; and (2) the robot’s torso is much narrower than long, offering limited lateral support.  
Furthermore, although \nickname demonstrates robustness on uneven terrains through domain randomization, its performance degrades on more challenging terrains such as stairs, primarily because the policy is trained only on flat ground environments.

\section{Acknowledgments}
\vspace{-0.2cm}
The work is partially supported by Google Deepmind with an unrestricted grant.


\bibliography{main} 

\begin{thebibliography}{60}
\providecommand{\natexlab}[1]{#1}
\providecommand{\url}[1]{\texttt{#1}}
\expandafter\ifx\csname urlstyle\endcsname\relax
  \providecommand{\doi}[1]{doi: #1}\else
  \providecommand{\doi}{doi: \begingroup \urlstyle{rm}\Url}\fi

\bibitem[Kumar et~al.(2021)Kumar, Fu, Pathak, and Malik]{kumar2021rma}
A.~Kumar, Z.~Fu, D.~Pathak, and J.~Malik.
\newblock Rma: Rapid motor adaptation for legged robots.
\newblock \emph{arXiv preprint arXiv:2107.04034}, 2021.

\bibitem[Agarwal et~al.(2023)Agarwal, Kumar, Malik, and Pathak]{agarwal2023legged}
A.~Agarwal, A.~Kumar, J.~Malik, and D.~Pathak.
\newblock Legged locomotion in challenging terrains using egocentric vision.
\newblock In \emph{Conference on robot learning}, pages 403--415. PMLR, 2023.

\bibitem[Margolis and Agrawal(2023)]{margolis2023walk}
G.~B. Margolis and P.~Agrawal.
\newblock Walk these ways: Tuning robot control for generalization with multiplicity of behavior.
\newblock In \emph{Conference on Robot Learning}, pages 22--31. PMLR, 2023.

\bibitem[Yang et~al.(2022)Yang, Zhang, Coumans, Tan, and Boots]{yang2022fast}
Y.~Yang, T.~Zhang, E.~Coumans, J.~Tan, and B.~Boots.
\newblock Fast and efficient locomotion via learned gait transitions.
\newblock In \emph{Conference on Robot Learning}, pages 773--783. PMLR, 2022.

\bibitem[Margolis et~al.(2024)Margolis, Yang, Paigwar, Chen, and Agrawal]{margolis2024rapid}
G.~B. Margolis, G.~Yang, K.~Paigwar, T.~Chen, and P.~Agrawal.
\newblock Rapid locomotion via reinforcement learning.
\newblock \emph{The International Journal of Robotics Research}, 43\penalty0 (4):\penalty0 572--587, 2024.

\bibitem[Cheng et~al.(2023)Cheng, Kumar, and Pathak]{cheng2023legs}
X.~Cheng, A.~Kumar, and D.~Pathak.
\newblock Legs as manipulator: Pushing quadrupedal agility beyond locomotion.
\newblock In \emph{2023 IEEE International Conference on Robotics and Automation (ICRA)}, pages 5106--5112. IEEE, 2023.

\bibitem[Ji et~al.(2023)Ji, Margolis, and Agrawal]{ji2023dribblebot}
Y.~Ji, G.~B. Margolis, and P.~Agrawal.
\newblock Dribblebot: Dynamic legged manipulation in the wild.
\newblock In \emph{2023 IEEE International Conference on Robotics and Automation (ICRA)}, pages 5155--5162. IEEE, 2023.

\bibitem[Huang et~al.(2023)Huang, Li, Xiang, Ni, Chi, Li, Yang, Peng, and Sreenath]{huang2023creating}
X.~Huang, Z.~Li, Y.~Xiang, Y.~Ni, Y.~Chi, Y.~Li, L.~Yang, X.~B. Peng, and K.~Sreenath.
\newblock Creating a dynamic quadrupedal robotic goalkeeper with reinforcement learning.
\newblock In \emph{2023 IEEE/RSJ International Conference on Intelligent Robots and Systems (IROS)}, pages 2715--2722. IEEE, 2023.

\bibitem[Fu et~al.(2023)Fu, Cheng, and Pathak]{fu2023deep}
Z.~Fu, X.~Cheng, and D.~Pathak.
\newblock Deep whole-body control: learning a unified policy for manipulation and locomotion.
\newblock In \emph{Conference on Robot Learning}, pages 138--149. PMLR, 2023.

\bibitem[Portela et~al.(2024)Portela, Margolis, Ji, and Agrawal]{portela2024learning}
T.~Portela, G.~B. Margolis, Y.~Ji, and P.~Agrawal.
\newblock Learning force control for legged manipulation.
\newblock In \emph{2024 IEEE International Conference on Robotics and Automation (ICRA)}, pages 15366--15372. IEEE, 2024.

\bibitem[Lin et~al.(2024)Lin, Liu, Yang, Niu, Yu, Zhang, Tan, Boots, and Zhao]{lin2024locoman}
C.~Lin, X.~Liu, Y.~Yang, Y.~Niu, W.~Yu, T.~Zhang, J.~Tan, B.~Boots, and D.~Zhao.
\newblock Locoman: Advancing versatile quadrupedal dexterity with lightweight loco-manipulators.
\newblock In \emph{2024 IEEE/RSJ International Conference on Intelligent Robots and Systems (IROS)}, pages 6877--6884. IEEE, 2024.

\bibitem[He et~al.(2025)He, Yuan, Zhou, Yang, Held, and Wang]{he2025visual}
X.~He, C.~Yuan, W.~Zhou, R.~Yang, D.~Held, and X.~Wang.
\newblock Visual manipulation with legs.
\newblock In \emph{Conference on Robot Learning}, pages 4218--4234. PMLR, 2025.

\bibitem[Valsecchi et~al.(2023)Valsecchi, Rudin, Nachtigall, Mayer, Tischhauser, and Hutter]{valsecchi2023barry}
G.~Valsecchi, N.~Rudin, L.~Nachtigall, K.~Mayer, F.~Tischhauser, and M.~Hutter.
\newblock Barry: a high-payload and agile quadruped robot.
\newblock \emph{IEEE Robotics and Automation Letters}, 8\penalty0 (11):\penalty0 6939--6946, 2023.

\bibitem[Amanzadeh et~al.(2024)Amanzadeh, Chunawala, Fawcett, Leonessa, and Hamed]{amanzadeh2024predictive}
L.~Amanzadeh, T.~Chunawala, R.~T. Fawcett, A.~Leonessa, and K.~A. Hamed.
\newblock Predictive control with indirect adaptive laws for payload transportation by quadrupedal robots.
\newblock \emph{IEEE Robotics and Automation Letters}, 2024.

\bibitem[Liu et~al.(2024)Liu, Chen, Cheng, Ji, Qiu, Yang, and Wang]{liu2024visual}
M.~Liu, Z.~Chen, X.~Cheng, Y.~Ji, R.-Z. Qiu, R.~Yang, and X.~Wang.
\newblock Visual whole-body control for legged loco-manipulation.
\newblock \emph{arXiv preprint arXiv:2403.16967}, 2024.

\bibitem[Yang et~al.(2022)Yang, Sue, Li, Yang, Shen, Chi, Rai, Zeng, and Sreenath]{yang2022collaborative}
C.~Yang, G.~N. Sue, Z.~Li, L.~Yang, H.~Shen, Y.~Chi, A.~Rai, J.~Zeng, and K.~Sreenath.
\newblock Collaborative navigation and manipulation of a cable-towed load by multiple quadrupedal robots.
\newblock \emph{IEEE Robotics and Automation Letters}, 7\penalty0 (4):\penalty0 10041--10048, 2022.

\bibitem[Jose and Zhang(2024)]{jose2024bilevel}
W.~J. Jose and H.~Zhang.
\newblock Bilevel learning for dual-quadruped collaborative transportation under kinematic and anisotropic velocity constraints.
\newblock \emph{arXiv preprint arXiv:2412.08644}, 2024.

\bibitem[Kim et~al.(2023)Kim, Fawcett, Kamidi, Ames, and Hamed]{kim2023layered}
J.~Kim, R.~T. Fawcett, V.~R. Kamidi, A.~D. Ames, and K.~A. Hamed.
\newblock Layered control for cooperative locomotion of two quadrupedal robots: Centralized and distributed approaches.
\newblock \emph{IEEE Transactions on Robotics}, 39\penalty0 (6):\penalty0 4728--4748, 2023.

\bibitem[Cheng et~al.(2024)Cheng, Shi, Agarwal, and Pathak]{cheng2024extreme}
X.~Cheng, K.~Shi, A.~Agarwal, and D.~Pathak.
\newblock Extreme parkour with legged robots.
\newblock In \emph{2024 IEEE International Conference on Robotics and Automation (ICRA)}, pages 11443--11450. IEEE, 2024.

\bibitem[Zhuang et~al.(2023)Zhuang, Fu, Wang, Atkeson, Schwertfeger, Finn, and Zhao]{zhuang2023robot}
Z.~Zhuang, Z.~Fu, J.~Wang, C.~Atkeson, S.~Schwertfeger, C.~Finn, and H.~Zhao.
\newblock Robot parkour learning.
\newblock \emph{arXiv preprint arXiv:2309.05665}, 2023.

\bibitem[Rudin et~al.(2022)Rudin, Hoeller, Reist, and Hutter]{rudin2022learning}
N.~Rudin, D.~Hoeller, P.~Reist, and M.~Hutter.
\newblock Learning to walk in minutes using massively parallel deep reinforcement learning.
\newblock In \emph{Conference on Robot Learning}, pages 91--100. PMLR, 2022.

\bibitem[Hoeller et~al.(2024)Hoeller, Rudin, Sako, and Hutter]{hoeller2024anymal}
D.~Hoeller, N.~Rudin, D.~Sako, and M.~Hutter.
\newblock Anymal parkour: Learning agile navigation for quadrupedal robots.
\newblock \emph{Science Robotics}, 9\penalty0 (88):\penalty0 eadi7566, 2024.

\bibitem[K{\"a}slin et~al.(2018)K{\"a}slin, Kolvenbach, Paez, Lika, and Hutter]{kaslin2018towards}
R.~K{\"a}slin, H.~Kolvenbach, L.~Paez, K.~Lika, and M.~Hutter.
\newblock Towards a passive adaptive planar foot with ground orientation and contact force sensing for legged robots.
\newblock In \emph{2018 IEEE/RSJ International Conference on Intelligent Robots and Systems (IROS)}, pages 2707--2714. IEEE, 2018.

\bibitem[Wu et~al.(2019)Wu, Huh, Sabin, Suresh, and Cutkosky]{wu2019tactile}
X.~A. Wu, T.~M. Huh, A.~Sabin, S.~A. Suresh, and M.~R. Cutkosky.
\newblock Tactile sensing and terrain-based gait control for small legged robots.
\newblock \emph{IEEE Transactions on Robotics}, 36\penalty0 (1):\penalty0 15--27, 2019.

\bibitem[Stone et~al.(2020)Stone, Lepora, and Barton]{stone2020walking}
E.~A. Stone, N.~F. Lepora, and D.~A. Barton.
\newblock Walking on tactip toes: A tactile sensing foot for walking robots.
\newblock In \emph{2020 IEEE/RSJ International Conference on Intelligent Robots and Systems (IROS)}, pages 9869--9875. IEEE, 2020.

\bibitem[Mudalige et~al.(2022)Mudalige, Nazarova, Babataev, Kopanev, Fedoseev, Cabrera, and Tsetserukou]{mudalige2022dogtouch}
N.~D.~W. Mudalige, E.~Nazarova, I.~Babataev, P.~Kopanev, A.~Fedoseev, M.~A. Cabrera, and D.~Tsetserukou.
\newblock Dogtouch: Cnn-based recognition of surface textures by quadruped robot with high density tactile sensors.
\newblock In \emph{2022 IEEE 95th Vehicular Technology Conference:(VTC2022-Spring)}, pages 1--5. IEEE, 2022.

\bibitem[Vangen et~al.(2023)Vangen, Barnwal, Olsen, and Alexis]{vangen2023terrain}
A.~Vangen, T.~Barnwal, J.~A. Olsen, and K.~Alexis.
\newblock Terrain recognition and contact force estimation through a sensorized paw for legged robots.
\newblock In \emph{2023 21st International Conference on Advanced Robotics (ICAR)}, pages 605--612. IEEE, 2023.

\bibitem[Shi et~al.(2024)Shi, Yao, Liu, Zhao, Zhu, and Jia]{shi2024foot}
G.~Shi, C.~Yao, X.~Liu, Y.~Zhao, Z.~Zhu, and Z.~Jia.
\newblock Foot vision: A vision-based multi-functional sensorized foot for quadruped robots.
\newblock \emph{IEEE Robotics and Automation Letters}, 2024.

\bibitem[Song et~al.(2024)Song, Li, Quan, Mu, Li, Zhao, Jin, Wu, Ding, and Zhang]{song2024tactid}
Z.~Song, C.~Li, Z.~Quan, S.~Mu, X.~Li, Z.~Zhao, W.~Jin, C.~Wu, W.~Ding, and X.-P. Zhang.
\newblock Tactid: High-performance visuo-tactile sensor-based terrain identification for legged robots.
\newblock \emph{IEEE Sensors Journal}, 2024.

\bibitem[Zhan et~al.(2023)Zhan, Cao, Chen, Guo, Gao, Luo, Guo, Zhou, and Gong]{zhan2023enable}
L.~Zhan, Y.~Cao, Q.~Chen, H.~Guo, J.~Gao, Y.~Luo, S.~Guo, G.~Zhou, and J.~Gong.
\newblock Enable natural tactile interaction for robot dog based on large-format distributed flexible pressure sensors.
\newblock In \emph{2023 IEEE International Conference on Robotics and Automation (ICRA)}, pages 12493--12499. IEEE, 2023.

\bibitem[Guo et~al.(2023)Guo, Zhan, Cao, Zheng, Zhou, and Gong]{guo2023touch}
S.~Guo, L.~Zhan, Y.~Cao, C.~Zheng, G.~Zhou, and J.~Gong.
\newblock Touch-and-heal: Data-driven affective computing in tactile interaction with robotic dog.
\newblock \emph{Proceedings of the ACM on Interactive, Mobile, Wearable and Ubiquitous Technologies}, 7\penalty0 (2):\penalty0 1--33, 2023.

\bibitem[Yu et~al.(2023)Yu, Han, Wang, Saxena, Xu, and Zhao]{yu2023mimictouch}
K.~Yu, Y.~Han, Q.~Wang, V.~Saxena, D.~Xu, and Y.~Zhao.
\newblock Mimictouch: Leveraging multi-modal human tactile demonstrations for contact-rich manipulation.
\newblock \emph{arXiv preprint arXiv:2310.16917}, 2023.

\bibitem[Guzey et~al.(2023)Guzey, Evans, Chintala, and Pinto]{guzey2023dexterity}
I.~Guzey, B.~Evans, S.~Chintala, and L.~Pinto.
\newblock Dexterity from touch: Self-supervised pre-training of tactile representations with robotic play.
\newblock \emph{arXiv preprint arXiv:2303.12076}, 2023.

\bibitem[Pattabiraman et~al.(2024)Pattabiraman, Cao, Haldar, Pinto, and Bhirangi]{pattabiraman2024learning}
V.~Pattabiraman, Y.~Cao, S.~Haldar, L.~Pinto, and R.~Bhirangi.
\newblock Learning precise, contact-rich manipulation through uncalibrated tactile skins.
\newblock \emph{arXiv preprint arXiv:2410.17246}, 2024.

\bibitem[Huang et~al.(2024)Huang, Wang, Yang, Luo, and Li]{huang20243d}
B.~Huang, Y.~Wang, X.~Yang, Y.~Luo, and Y.~Li.
\newblock 3d-vitac: Learning fine-grained manipulation with visuo-tactile sensing.
\newblock \emph{arXiv preprint arXiv:2410.24091}, 2024.

\bibitem[Xue et~al.(2025)Xue, Ren, Chen, Zhang, Fang, Gu, Xu, and Lu]{xue2025reactive}
H.~Xue, J.~Ren, W.~Chen, G.~Zhang, Y.~Fang, G.~Gu, H.~Xu, and C.~Lu.
\newblock Reactive diffusion policy: Slow-fast visual-tactile policy learning for contact-rich manipulation.
\newblock \emph{arXiv preprint arXiv:2503.02881}, 2025.

\bibitem[Zhang et~al.(2025)Zhang, Hu, Yuan, and Xu]{zhang2025doglove}
H.~Zhang, S.~Hu, Z.~Yuan, and H.~Xu.
\newblock Doglove: Dexterous manipulation with a low-cost open-source haptic force feedback glove.
\newblock \emph{arXiv preprint arXiv:2502.07730}, 2025.

\bibitem[Yin et~al.(2023)Yin, Huang, Qin, Chen, and Wang]{yin2023rotating}
Z.-H. Yin, B.~Huang, Y.~Qin, Q.~Chen, and X.~Wang.
\newblock Rotating without seeing: Towards in-hand dexterity through touch.
\newblock \emph{arXiv preprint arXiv:2303.10880}, 2023.

\bibitem[Yuan et~al.(2024)Yuan, Che, Qin, Huang, Yin, Lee, Wu, Lim, and Wang]{yuan2024robot}
Y.~Yuan, H.~Che, Y.~Qin, B.~Huang, Z.-H. Yin, K.-W. Lee, Y.~Wu, S.-C. Lim, and X.~Wang.
\newblock Robot synesthesia: In-hand manipulation with visuotactile sensing.
\newblock In \emph{2024 IEEE International Conference on Robotics and Automation (ICRA)}, pages 6558--6565. IEEE, 2024.

\bibitem[Xue et~al.(2024)Xue, Zhang, Cheng, He, Ju, Lin, Zhang, and Xu]{xue2024arraybot}
Z.~Xue, H.~Zhang, J.~Cheng, Z.~He, Y.~Ju, C.~Lin, G.~Zhang, and H.~Xu.
\newblock Arraybot: Reinforcement learning for generalizable distributed manipulation through touch.
\newblock In \emph{2024 IEEE International Conference on Robotics and Automation (ICRA)}, pages 16744--16751. IEEE, 2024.

\bibitem[Yin et~al.(2024)Yin, Qi, Malik, Pikul, Yim, and Hellebrekers]{yin2024learning}
J.~Yin, H.~Qi, J.~Malik, J.~Pikul, M.~Yim, and T.~Hellebrekers.
\newblock Learning in-hand translation using tactile skin with shear and normal force sensing.
\newblock \emph{arXiv preprint arXiv:2407.07885}, 2024.

\bibitem[Narang et~al.(2020)Narang, Van~Wyk, Mousavian, and Fox]{narang2020interpreting}
Y.~S. Narang, K.~Van~Wyk, A.~Mousavian, and D.~Fox.
\newblock Interpreting and predicting tactile signals via a physics-based and data-driven framework.
\newblock \emph{arXiv preprint arXiv:2006.03777}, 2020.

\bibitem[Narang et~al.(2021)Narang, Sundaralingam, Macklin, Mousavian, and Fox]{narang2021sim}
Y.~Narang, B.~Sundaralingam, M.~Macklin, A.~Mousavian, and D.~Fox.
\newblock Sim-to-real for robotic tactile sensing via physics-based simulation and learned latent projections.
\newblock In \emph{2021 IEEE International Conference on Robotics and Automation (ICRA)}, pages 6444--6451. IEEE, 2021.

\bibitem[Kappassov et~al.(2020)Kappassov, Corrales-Ramon, and Perdereau]{kappassov2020simulation}
Z.~Kappassov, J.-A. Corrales-Ramon, and V.~Perdereau.
\newblock Simulation of tactile sensing arrays for physical interaction tasks.
\newblock In \emph{2020 IEEE/ASME International Conference on Advanced Intelligent Mechatronics (AIM)}, pages 196--201. IEEE, 2020.

\bibitem[Kasolowsky and B{\"a}uml(2024)]{kasolowsky2024fine}
U.~Kasolowsky and B.~B{\"a}uml.
\newblock Fine manipulation using a tactile skin: Learning in simulation and sim-to-real transfer.
\newblock In \emph{2024 IEEE/RSJ International Conference on Intelligent Robots and Systems (IROS)}, pages 13120--13127. IEEE, 2024.

\bibitem[Leins et~al.(2025)Leins, Patzelt, and Haschke]{leins2025hydroelastictouch}
D.~P. Leins, F.~Patzelt, and R.~Haschke.
\newblock Hydroelastictouch: Simulation of tactile sensors with hydroelastic contact surfaces.
\newblock \emph{arXiv preprint arXiv:2501.08077}, 2025.

\bibitem[Ding et~al.(2021)Ding, Tsai, Lee, and Huang]{ding2021sim}
Z.~Ding, Y.-Y. Tsai, W.~W. Lee, and B.~Huang.
\newblock Sim-to-real transfer for robotic manipulation with tactile sensory.
\newblock In \emph{2021 IEEE/RSJ International Conference on Intelligent Robots and Systems (IROS)}, pages 6778--6785. IEEE, 2021.

\bibitem[Yang et~al.(2023)Yang, Huang, Li, Tsai, Lee, Song, and Pan]{yang2023tacgnn}
L.~Yang, B.~Huang, Q.~Li, Y.-Y. Tsai, W.~W. Lee, C.~Song, and J.~Pan.
\newblock Tacgnn: Learning tactile-based in-hand manipulation with a blind robot using hierarchical graph neural network.
\newblock \emph{IEEE Robotics and Automation Letters}, 8\penalty0 (6):\penalty0 3605--3612, 2023.

\bibitem[Schulman et~al.(2017)Schulman, Wolski, Dhariwal, Radford, and Klimov]{schulman2017proximal}
J.~Schulman, F.~Wolski, P.~Dhariwal, A.~Radford, and O.~Klimov.
\newblock Proximal policy optimization algorithms.
\newblock \emph{arXiv preprint arXiv:1707.06347}, 2017.

\bibitem[Ross et~al.(2011)Ross, Gordon, and Bagnell]{ross2011reduction}
S.~Ross, G.~Gordon, and D.~Bagnell.
\newblock A reduction of imitation learning and structured prediction to no-regret online learning.
\newblock In \emph{Proceedings of the fourteenth international conference on artificial intelligence and statistics}, pages 627--635. JMLR Workshop and Conference Proceedings, 2011.

\bibitem[Sundaram et~al.(2019)Sundaram, Kellnhofer, Li, Zhu, Torralba, and Matusik]{sundaram2019learning}
S.~Sundaram, P.~Kellnhofer, Y.~Li, J.-Y. Zhu, A.~Torralba, and W.~Matusik.
\newblock Learning the signatures of the human grasp using a scalable tactile glove.
\newblock \emph{Nature}, 569\penalty0 (7758):\penalty0 698--702, 2019.

\bibitem[Mittal et~al.(2023)Mittal, Yu, Yu, Liu, Rudin, Hoeller, Yuan, Singh, Guo, Mazhar, et~al.]{mittal2023orbit}
M.~Mittal, C.~Yu, Q.~Yu, J.~Liu, N.~Rudin, D.~Hoeller, J.~L. Yuan, R.~Singh, Y.~Guo, H.~Mazhar, et~al.
\newblock Orbit: A unified simulation framework for interactive robot learning environments.
\newblock \emph{IEEE Robotics and Automation Letters}, 8\penalty0 (6):\penalty0 3740--3747, 2023.

\bibitem[Yang et~al.(2024)Yang, Shi, Lin, Meng, Scalise, Castro, Yu, Zhang, Zhao, Tan, et~al.]{yang2024agile}
Y.~Yang, G.~Shi, C.~Lin, X.~Meng, R.~Scalise, M.~G. Castro, W.~Yu, T.~Zhang, D.~Zhao, J.~Tan, et~al.
\newblock Agile continuous jumping in discontinuous terrains.
\newblock \emph{arXiv preprint arXiv:2409.10923}, 2024.

\bibitem[Patterson et~al.(2008)Patterson, Parafianowicz, Danells, Closson, Verrier, Staines, Black, and McIlroy]{patterson2008gait}
K.~K. Patterson, I.~Parafianowicz, C.~J. Danells, V.~Closson, M.~C. Verrier, W.~R. Staines, S.~E. Black, and W.~E. McIlroy.
\newblock Gait asymmetry in community-ambulating stroke survivors.
\newblock \emph{Archives of physical medicine and rehabilitation}, 89\penalty0 (2):\penalty0 304--310, 2008.

\bibitem[Yu et~al.(2018)Yu, Turk, and Liu]{yu2018learning}
W.~Yu, G.~Turk, and C.~K. Liu.
\newblock Learning symmetric and low-energy locomotion.
\newblock \emph{ACM Transactions on Graphics (TOG)}, 37\penalty0 (4):\penalty0 1--12, 2018.

\bibitem[Abdolhosseini et~al.(2019)Abdolhosseini, Ling, Xie, Peng, and Van~de Panne]{abdolhosseini2019learning}
F.~Abdolhosseini, H.~Y. Ling, Z.~Xie, X.~B. Peng, and M.~Van~de Panne.
\newblock On learning symmetric locomotion.
\newblock In \emph{Proceedings of the 12th ACM SIGGRAPH Conference on Motion, Interaction and Games}, pages 1--10, 2019.

\bibitem[Su et~al.(2024)Su, Huang, Ordo{\~n}ez-Apraez, Li, Li, Liao, Turrisi, Pontil, Semini, Wu, et~al.]{su2024leveraging}
Z.~Su, X.~Huang, D.~Ordo{\~n}ez-Apraez, Y.~Li, Z.~Li, Q.~Liao, G.~Turrisi, M.~Pontil, C.~Semini, Y.~Wu, et~al.
\newblock Leveraging symmetry in rl-based legged locomotion control.
\newblock In \emph{2024 IEEE/RSJ International Conference on Intelligent Robots and Systems (IROS)}, pages 6899--6906. IEEE, 2024.

\bibitem[Mittal et~al.(2024)Mittal, Rudin, Klemm, Allshire, and Hutter]{mittal2024symmetry}
M.~Mittal, N.~Rudin, V.~Klemm, A.~Allshire, and M.~Hutter.
\newblock Symmetry considerations for learning task symmetric robot policies.
\newblock In \emph{2024 IEEE International Conference on Robotics and Automation (ICRA)}, pages 7433--7439. IEEE, 2024.

\bibitem[Hutter et~al.(2016)Hutter, Gehring, Jud, Lauber, Bellicoso, Tsounis, Hwangbo, Bodie, Fankhauser, Bloesch, et~al.]{hutter2016anymal}
M.~Hutter, C.~Gehring, D.~Jud, A.~Lauber, C.~D. Bellicoso, V.~Tsounis, J.~Hwangbo, K.~Bodie, P.~Fankhauser, M.~Bloesch, et~al.
\newblock Anymal-a highly mobile and dynamic quadrupedal robot.
\newblock In \emph{2016 IEEE/RSJ international conference on intelligent robots and systems (IROS)}, pages 38--44. IEEE, 2016.

\bibitem[Bledt et~al.(2018)Bledt, Powell, Katz, Di~Carlo, Wensing, and Kim]{bledt2018cheetah}
G.~Bledt, M.~J. Powell, B.~Katz, J.~Di~Carlo, P.~M. Wensing, and S.~Kim.
\newblock Mit cheetah 3: Design and control of a robust, dynamic quadruped robot.
\newblock In \emph{2018 IEEE/RSJ International Conference on Intelligent Robots and Systems (IROS)}, pages 2245--2252. IEEE, 2018.

\end{thebibliography}

\newpage
\appendix
\section{Related Work on Symmetric Locomotion Learning}
Symmetry is a fundamental property of healthy gaits, contributing to locomotion stability, adaptability, and reduced risk of falling~\cite{patterson2008gait}. Various methods have been proposed to enforce gait symmetry in reinforcement learning-based locomotion policies and have demonstrated notable effectiveness~\cite{yu2018learning, abdolhosseini2019learning, su2024leveraging, mittal2024symmetry}. The common principle behind these approaches is that policies should produce symmetric \emph{actions} when observing symmetric \emph{states}, where the definition of action symmetry is derived from the kinematic symmetry of the robot. To achieve this, previous work introduces mechanisms such as mirror symmetry losses~\cite{yu2018learning} and symmetry-based data augmentation~\cite{su2024leveraging, mittal2024symmetry}, enforcing them over the entire \emph{state–action} space.

Despite their effectiveness, these approaches have several limitations:
\begin{enumerate}
    \item \textbf{Robot-specific assumptions.} The definition of mirror symmetry depends heavily on robot morphology. For example, ETH Anymal~\cite{hutter2016anymal} is symmetric both left-right and front-rear, while MIT Cheetah 3 robots~\cite{bledt2018cheetah} and Unitree Go1 are only left-right symmetric. This requires a careful and manual specification of symmetry mappings for each new platform.
    \item \textbf{Mismatch between kinematic and physical symmetry.} These methods often assume that robot dynamics and sensor configurations share the same symmetry as kinematics. In practice, real robots rarely satisfy this assumption due to asymmetries in hardware, sensors, or payload.
    \item \textbf{High-dimensional state challenges.} Enforcing symmetry over the entire state–action space becomes increasingly difficult as observation spaces grow, particularly for policies that incorporate high-dimensional sensory inputs such as images.
    \item \textbf{Neglect of temporal symmetry.} These methods promote symmetry only in kinematic trajectories, overlooking temporal aspects such as balanced swing and stance durations across legs.
\end{enumerate}

Our key insight is that robots should exhibit symmetric gait behaviors when task commands (e.g., velocity commands) remain constant, and that this requirement can be relaxed under degraded task performance. For example, when the robot or its carried object experiences a sudden external disturbance, it may need to adjust step timing asymmetrically to maintain balance. In our formulation, gait symmetry is defined as equal swing or stance durations within each pair of legs (e.g., diagonal pairs in trotting gaits).
Based on this insight, we propose a \emph{symmetricity function} that evaluates symmetry directly in terms of \emph{gait behavior (swing–stance phases)} rather than kinematic states or actions. This approach is independent of robot morphology, sensor configuration, and environmental conditions, making it broadly applicable across tasks and platforms. Moreover, by integrating this symmetricity function into the reward, we directly encourage gait-level symmetry through policy gradients, without introducing additional loss terms or complicating the learning paradigm.

For clarity, the key comparison metrics with prior methods are summarized in Table~\ref{tab:symmetric_methods}.
\vspace{-0.15cm}
\begin{table}[h]
    \centering
    \caption{Comparison of methods for symmetric locomotion learning.}
    \vspace{0.15cm}
    \begin{tabular}{lcc}
    \toprule
        & Prior Works~\cite{yu2018learning, abdolhosseini2019learning, su2024leveraging, mittal2024symmetry} & \nickname \\
    \midrule
        Symmetry Basis & Robot Kinematics & Gait Definition \\
        Symmetry Condition & States or Observations & Commands and Task Performance \\
        Symmetry Target & Whole-Body Actions & Foot Contact Patterns \\
        Enforcement & Auxiliary Loss or Augmentation & Reward Function \\
    \bottomrule
    \end{tabular}
    \label{tab:symmetric_methods}
\end{table}

\newpage
\section{Details for Distributed Tactile Sensors}
\label{sec:app_tactile}
\subsection{Principle and Fabrication}
As introduced in Section~\ref{sec:tactile_sensing}, our tactile sensor leverages the principle of piezoresistive sensing, where the electrical resistance of certain materials changes under mechanical deformation. Similar to tactile gloves~\cite{sundaram2019learning} and tactile grippers~\cite{huang20243d}, we use Velostat (Linqstat) as the piezoresistive material.

However, our conductive layer design differs significantly from them~\cite{sundaram2019learning,huang20243d}, leading to substantial improvements:
\emph{(a) High Sensitivity.} Previous designs rely on conductive threads, which make only minimal contact with the piezoresistive layer, thereby limiting sensitivity. In contrast, we use a thin Faraday conductive fabric that provides a much larger contact area. This design greatly enhances sensitivity, allowing the sensor to reliably detect even a lightweight object such as a penny without external pressing.
\emph{(b) Scalable Fabrication.} Conductive threads require manual placement, which is time-consuming, imprecise, and prone to air bubbles within the adhesive layers.
By contrast, our method employs laser cutting of the conductive fabric, enabling efficient, precise, and highly scalable fabrication simply by stacking pre-cut layers.

\begin{figure}[h]
  \centering
  \includegraphics[width=\linewidth]{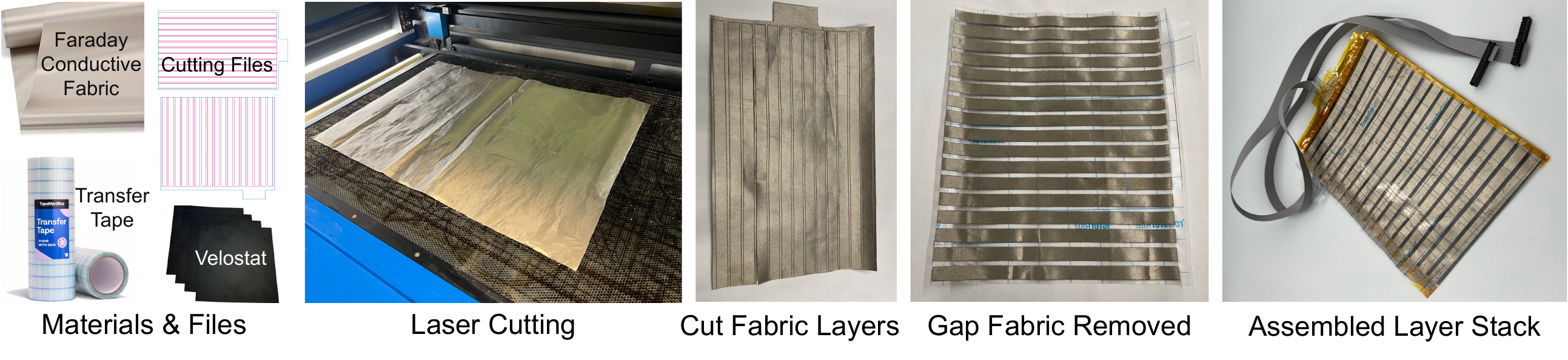}
  \caption{Materials, cutting files, and fabrication procedures for building the tactile sensors. The illustrated sensor shape is for demonstration and is differ from the final version used in \nickname.}
  \label{fig:frabricartion}
\end{figure}

As shown in Figure~\ref{fig:frabricartion}, the main components of the tactile sensor are:
(1) Velostat, which serves as the piezoresistive material;
(2) Faraday conductive fabric, which interfaces with Velostat;
(3) Transfer tape, which fastens and encapsulates the layers.

The fabrication process consists of the following steps:
\begin{enumerate}
    \item Attach the conductive fabric onto the transfer tape.
    \item Laser cut the rows or columns on the conductive fabric, carefully tuning laser power and speed to avoid cutting through the transfer tape. Fiber laser would be ideal since it doesn't cut through the transfer tape.
    \item Laser cut the outer shape of the sensor on both the conductive fabric and the transfer tape.
    \item Remove the excess conductive fabric, leaving only the patterned rows or columns adhered to the transfer tape.
    \item Attach external cables to one end of the rows and columns for readout.
    \item Laser cut the Velostat layer and stack it with the two conductive layers prepared above.
\end{enumerate}
For additional resources on piezoresistive tactile sensors, we refer readers to the open-source \href{https://yyueluo.com/tactile-skin-tool/}{tactile-skin toolkit}.

\subsection{Expanded Collision Model Calibration}

\begin{figure}[h]
  \centering
  \includegraphics[width=0.95\linewidth]{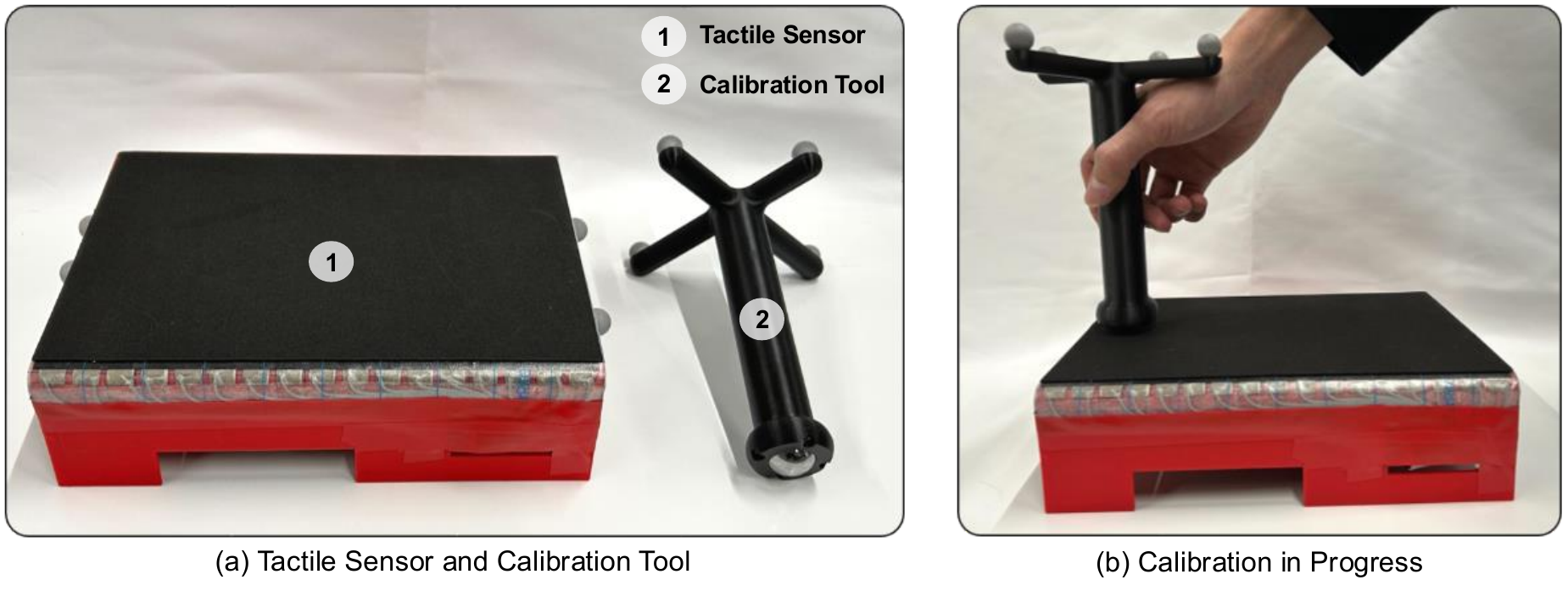}
  \caption{Hardware set-up for expanded collision model calibration.}
  \label{fig:simulation_cali}
\end{figure}

\begin{figure}[h]
  \centering
  \includegraphics[width=0.95\linewidth]{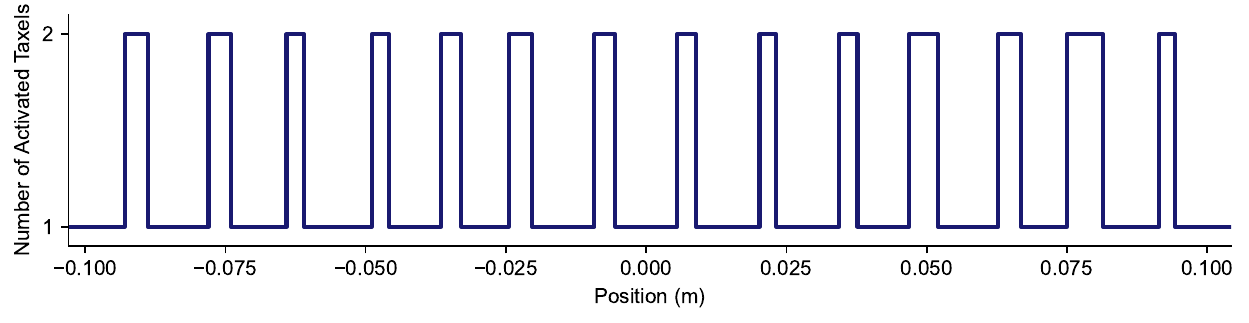}
  \caption{The ball activates one or two taxels during calibration.}
  \label{fig:simulation_cali_plot}
\end{figure}
We design a calibration tool (Figure~\ref{fig:simulation_cali}) with a metal ball mounted at its end to calibrate the spatial dimensions of the expanded tactile model. During calibration, we press the tool onto the tactile sensor and slide it from left to right, recording both the tactile signals and the ball's position using a motion capture (MoCap) system. Each trial targets a single row or column of taxels. Based on the resulting binary activation map, we determine whether the ball activates a single taxel or two adjacent taxels. We then plot the number of activated taxels within the target row or column (Figure~\ref{fig:simulation_cali_plot}) and compute the mean width of the overlap regions. The tactile sensor is covered with a soft black foam layer, which protects the surface and reduces signal loss in large contact areas.

\subsection{Latency Calibration}
\begin{figure}[h]
  \centering
  \begin{minipage}[t]{0.27\linewidth}
    \centering
    \includegraphics[width=\linewidth]{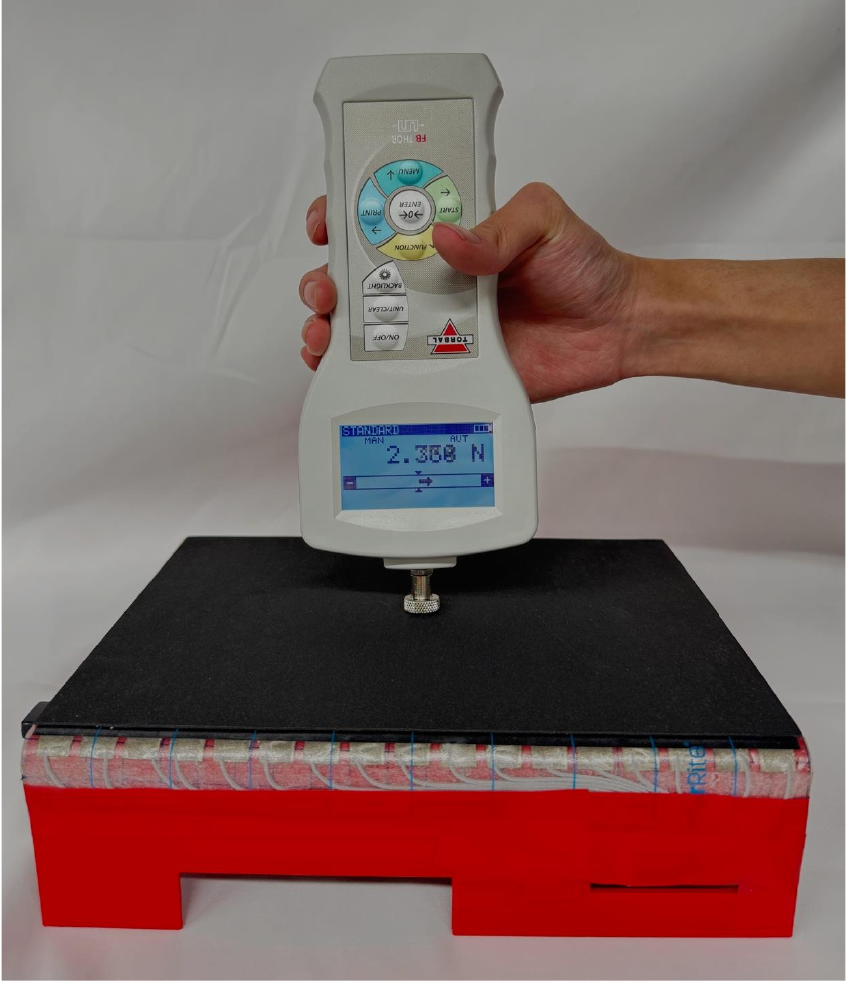}
  \end{minipage}%
  \hspace{1em}
  \begin{minipage}[t]{0.63\linewidth}
    \centering
    \includegraphics[width=\linewidth]{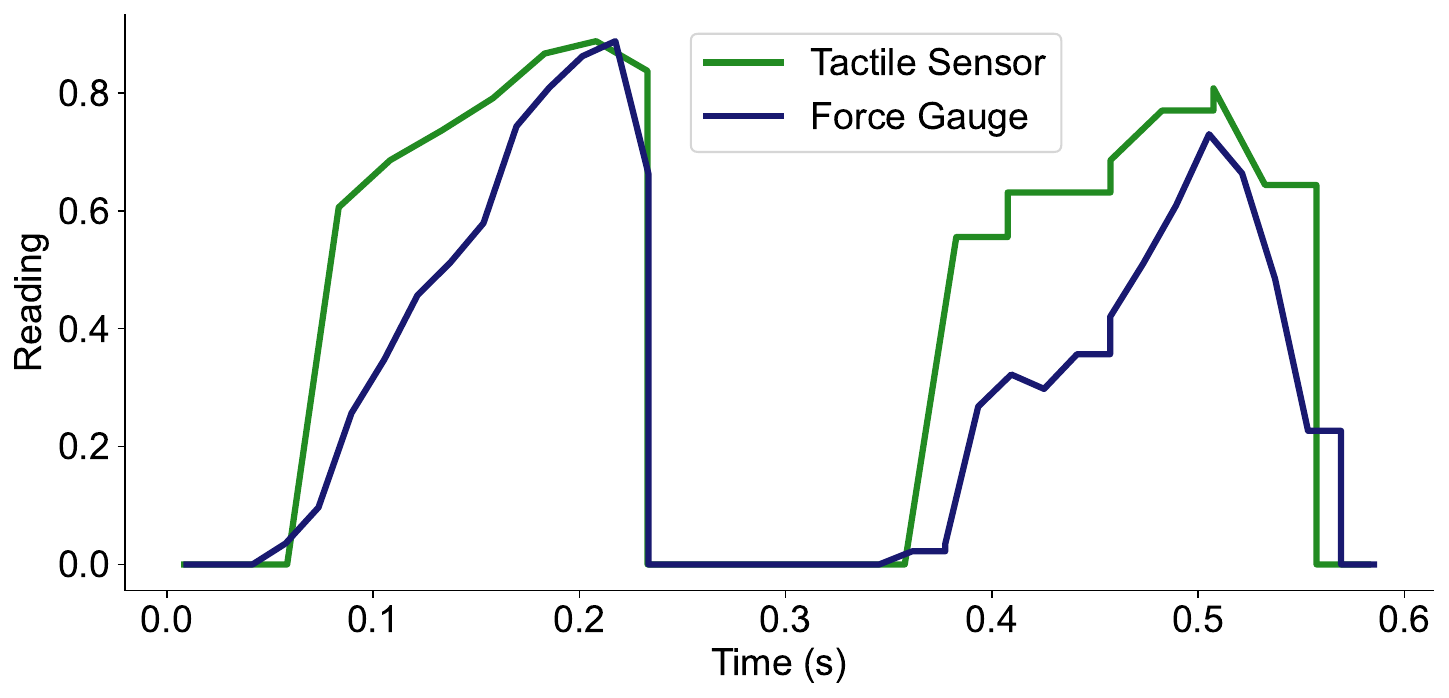}
  \end{minipage}
  \caption{\textbf{Left:} Hardware set-up for sensor latency calibration. \textbf{Right:} Plot of readings from the tactile sensor and the force gauge.}
  \label{fig:latency_cali}
\end{figure}

We use a force gauge to press on our distributed tactile sensor and record the readings from both devices. As shown in Figure~\ref{fig:latency_cali}, the tactile sensor readings exhibit a delay of approximately $0.02,\mathrm{s}$ compared to the force gauge. The force gauge, operating at $80\mathrm{Hz}$ with negligible latency, serves as a reliable reference. Based on this comparison, we estimate the tactile sensor's latency to be in the range of $0.02$–$0.04,\mathrm{s}$, which is relatively low compared to that of a depth camera. Using a higher-frequency force gauge would allow for more precise latency calibration, and we plan to adopt such equipment in future work.

\section{Details for Tactile Policy Training}
To begin with, we define the notation in Table~\ref{tab:notation}.

\begin{table*}[h]
    \centering
    \caption{Notations.}
    \vspace{-0.15cm}
    \renewcommand{\arraystretch}{1.1}
    \begin{tabular}{clcc}
    \toprule
        & Term & Notation & Unit \\
    \midrule
        \multicolumn{1}{c}{\multirow{2}{*}{Reference Frame}}
         & Robot frame & $(\cdot)^\text{r}$ & -\\
         & World frame & $(\cdot)^\text{w}$ & -\\
    \midrule
        \multicolumn{1}{c}{\multirow{5}{*}{Object}}
         & Position & $\boldsymbol{p}_\text{o}^\text{r}$ & $m$\\
         & Linear velocity & $\boldsymbol{v}_\text{o}^\text{r}$ & $m/s$\\
         & Quaternion & $\boldsymbol{Q}_\text{o}^\text{r}$ & -\\
         & Euler angle & $\boldsymbol{\theta}_\text{o}^\text{r}$ & $rad$\\
         & Angular velocity & $\boldsymbol{w}_\text{o}^\text{r}$ & $rad/s$\\
    \midrule
        \multicolumn{1}{c}{\multirow{5}{*}{Robot Torso}}
         & Position  & $\boldsymbol{p}^\text{w}$ & $m$\\
         & Linear velocity  & $\boldsymbol{v}^\text{r}$ & $m/s$\\
         & Quaternion  & $\boldsymbol{Q}^\text{w}$ & -\\
         & Euler angle  & $\boldsymbol{\theta}^\text{w}$ & $rad$\\
         & Angular velocity  & $\boldsymbol{w}^\text{r}$ & $rad/s$\\
         \cmidrule(lr){2-4}
        \multicolumn{1}{c}{\multirow{3}{*}{Robot Joint}}
         & Joint position  & $\boldsymbol{q}$ & $rad$\\
         & Joint velocity  & $\dot{\boldsymbol{q}}$ & $rad/s$\\
         & Joint acceleration  & $\ddot{\boldsymbol{q}}$ & $rad/s^2$\\
         \cmidrule(lr){2-4}
        \multicolumn{1}{c}{\multirow{5}{*}{Robot Foot}}
         & Foot position  & $\boldsymbol{p}_\text{f}^\text{w}$ & $m$\\
         & Foot velocity  & $\boldsymbol{v}_\text{f}^\text{w}$ & $m/s$\\
         & Foot contact force  & $\boldsymbol{F}_\text{f}^\text{w}$ & $N$\\
         & Foot current air time  & $t_\text{curr}$ & $s$\\
         & Foot previous air time  & $t_\text{pre}$ & $s$\\
    \bottomrule
    \end{tabular}
    \label{tab:notation}
\end{table*}

\subsection{Basic Configurations in Simulation}
\begin{figure}[h]
  \centering
  \includegraphics[width=0.9\linewidth]{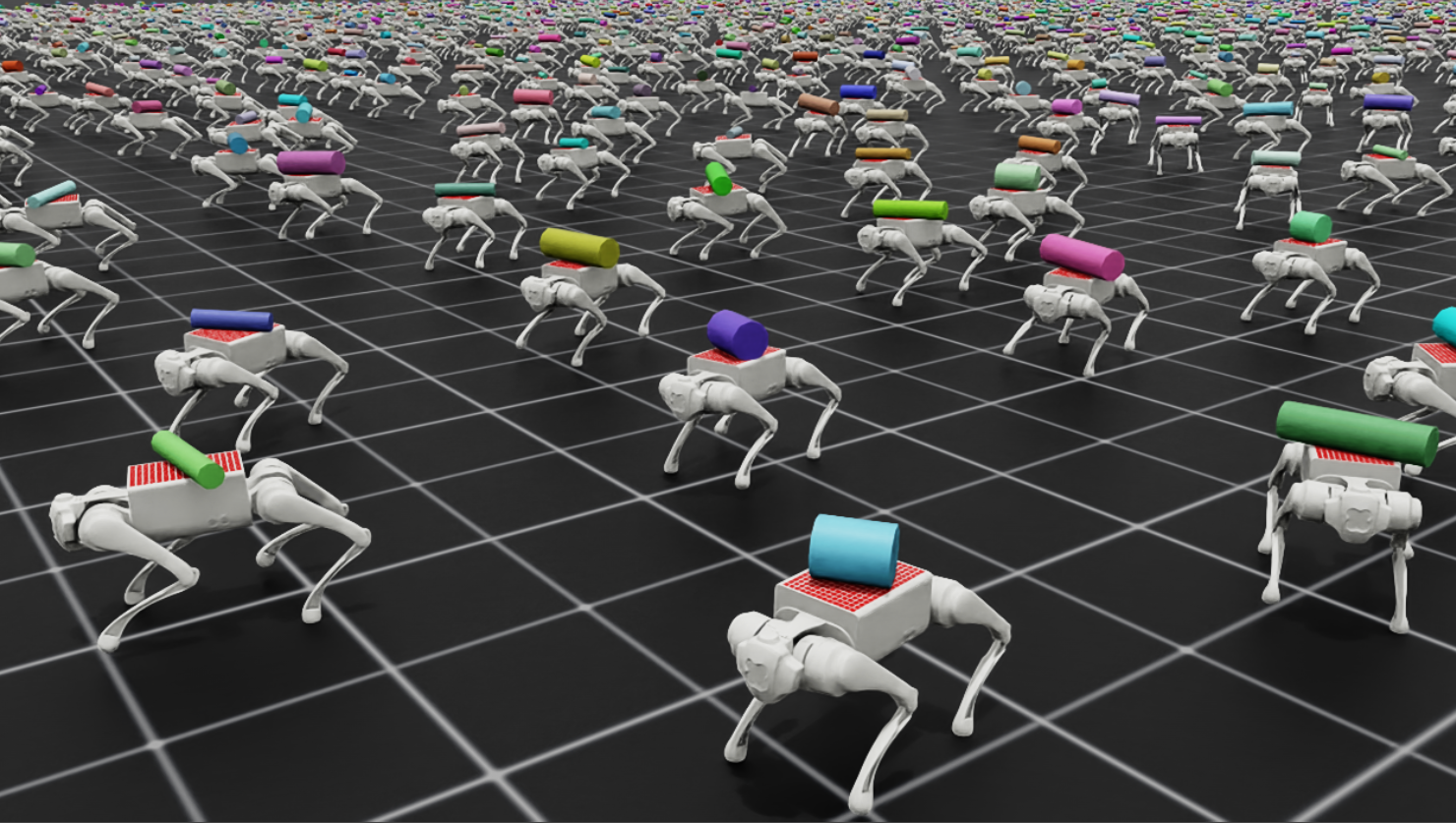}
  \caption{Simulation environment for teacher policy training in IsaacLab.}
  \label{fig:teacher-isaaclab}
\end{figure}
\textbf{Training Environment.}
As Figure~\ref{fig:teacher-isaaclab} illustrates, we train the teacher policy in IsaacLab~\cite{mittal2023orbit} using 4096 parallel simulation environments. The robot includes collision models for both the back shell and the tactile sensors; however, force tracking for the tactile sensors is disabled during this phase to speed up the contact computation, as the teacher policy does not rely on tactile signals. To improve robustness, we apply domain randomization to the environment, using the terms listed in Table~\ref{tab:randomization}.

\begin{table*}[h]
    \centering
    \caption{Domain randomizations and their ranges during teacher policy training.}
    \vspace{-0.15cm}
    \begin{tabular}{llccc}
    \toprule
        & Randomized Term & Randomization Range & Unit \\
    \midrule
         \multicolumn{1}{l}{\multirow{5}{*}{Object}}
         & \multicolumn{1}{l}{\multirow{3}{*}{Initial Pose}}
         & $x\in[-0.05, 0.05]$ & $m$ \\
         & & $y\in[-0.04, 0.04]$ & $m$ \\
         & & $yaw\in[-30, 30]$ & \degree \\
         \cmidrule(lr){2-4}
         & Mass & $[0.5, 2.5]$ & $kg$ \\
         & Friction Coefficient & $[0.3, 1.0]$ & - \\
    \midrule
         \multicolumn{1}{l}{\multirow{5}{*}{Robot}}
         & Trunk Mass & $[4.2, 6.2]$ & $kg$ \\
         & Trunk Friction Coefficient & $[0.3, 1.0]$ & - \\
         & Foot Friction Coefficient & $[0.6, 1.5]$ & - \\
         & Initial Delta Joint Position & $[-0.03, 0.03]$ & $rad$ \\
         & Initial Joint Velocity & $[-0.1, 0.1]$ & $rad/s$ \\
    \midrule
         \multicolumn{1}{l}{\multirow{6}{*}{Pushing}}
         & \multicolumn{1}{l}{\multirow{3}{*}{Object}}
         & $v_x\in[-0.3, 0.3]$ & $m/s$ \\
         & & $v_y\in[-0.3, 0.3]$ & $m/s$ \\
         & & $v_z\in[-0.2, 0.2]$ & $m/s$ \\
         \cmidrule(lr){2-4}
         & \multicolumn{1}{l}{\multirow{3}{*}{Robot Trunk}}
         & $v_x\in[-0.4, 0.4]$ & $m/s$ \\
         & & $v_y\in[-0.3, 0.3]$ & $m/s$ \\
         & & $v_z\in[-0.1, 0.1]$ & $m/s$ \\
    \bottomrule
    \end{tabular}
    \label{tab:randomization}
\end{table*}

\textbf{Policy Network.}
The teacher policy (actor) is implemented as a multi-layer perceptron (MLP) that takes as input the state sequence $o_{(t-H) \dots t} \in \mathbb{R}^{340}$. The network consists of four fully connected layers with hidden dimensions of $512$, $256$, and $128$, each followed by an ELU activation.  It outputs an action vector $a_t \in \mathbb{R}^{12}$, representing the target delta joint positions. This output is clipped to the range $[-100, 100]$ for stable training and scaled by a factor $\alpha^{\text{action}} = 0.25$, producing the target joint positions as $q_t^{\text{target}} = \alpha^{\text{action}} \cdot a_t + q^{\text{default}}$, where $q^{\text{default}}$ is the default joint positions of the robot. A PD controller with gains $K_p = 25$ and $K_d = 0.5$ is then used to track the target joint positions $q_t^{target}$.

\textbf{Post-Processing for Observations.}
To improve robustness and narrow the domain gap for the tactile student policy which relies on partial observations, we add noise to the actor's input during training. Each processed observation term is computed as $o_i^{input} = \alpha_i^{input}\cdot(o_i^{raw} + o_i^{noise})$, where $o_i^{raw}$ is the $i$-th observation term from simulation, $o_i^{noise}$ is the added noise, $\alpha_i^{input}$ is a scaling factor, and $o_i^{input}$ is the processed observation term fed into the policy.
Notably, for the object orientation, noise is first added to the Euler angles before converting them to quaternions which are finally used as the input to the policy.
The details of each observation term, including noise ranges and scaling factors, are summarized in Table~\ref{tab:noise}.
\begin{table*}[h]
    \centering
    \caption{Noise and scaling applied to each observation term during teacher policy training.}
    \vspace{-0.15cm}
    \begin{tabular}{llccc}
    \toprule
        & Observation Term & Noise Range & Scale & Unit \\
    \midrule
         \multicolumn{1}{l}{\multirow{4}{*}{Object State}}
         & Position & $[-0.01, 0.01]$ & 1.0 & $m$ \\
         & Linear Velocity & $[-0.2, 0.2]$ & 0.5 & $m/s$ \\
         & Orientation & $[-0.05, 0.05]$ & 1.0 & $rad$ \\
         & Angular Velocity & $[-0.2, 0.2]$ & 0.25 & $rad/s$ \\
    \midrule
         \multicolumn{1}{l}{\multirow{4}{*}{Robot Proprioception}}
         & Projected Gravity Vector & $[-0.05, 0.05]$ & 1.0 & - \\
         & Base Angular Velocity & $[-0.2, 0.2]$ & 0.25 & $rad/s$ \\
         & Joint Position & $[-0.01, 0.01]$ & 1.0 & $rad$ \\
         & Joint Position & $[-1.5, 1.5]$ & 0.05 & $rad/s$ \\
    \midrule
        \multicolumn{1}{l}{\multirow{2}{*}{Others}}
         &Velocity Command & - & 1.0 & $m/s$ or $rad/s$ \\
         &Last Action & - & 1.0 & $rad$ \\
        \bottomrule
    \end{tabular}
    \label{tab:noise}
\end{table*}

\subsection{Curriculum of Velocity Commands}
\label{sec:vel_curriculum}
To encourage accurate and stable tracking, we adopt a curriculum that gradually expands the range of velocity commands as the policy achieves predefined thresholds in tracking performance and survival rate.
At the beginning of training, both linear and angular velocity commands are constrained to small but non-zero ranges.
These ranges are then symmetrically expanded step by step, with linear and angular components progressing independently until they reach their predefined maximum values. To maintain balanced development, we further restrict how far one curriculum can advance relative to the other: if the number of completed expansions for linear and angular ranges differs by more than a specified margin, further expansion is paused until the lagging dimension catches up.

\subsection{Schedule of Zero Velocity Commands}
\label{sec:zero_vel_schedule}
As discussed in Section~\ref{sec:teacher}, we set the velocity commands to zero during the first $k$ steps of each episode ($k=50$ in our implementation) to ensure stable object placement and enable successful distillation from the teacher policy to the student policy with partial observations such as tactile sensing. This warm-up period allows the object to stabilize on the robot’s back before nonzero motion commands are applied.
However, we observe that applying zero commands from the very beginning severely limits exploration, especially in the early phase of PPO training. Since most episodes terminate before reaching the $k$-th step, the policy is encouraged to maintain ground contact with all feet under zero commands, leading to premature convergence to relatively static behaviors rather than exploring long-swinging dynamic gaits.

To address this, we introduce a curriculum on the initial zero-command steps. At the beginning of training, we set the initial step length with zero commands to zero, which encourages a broader exploration of locomotion gaits. Once the command ranges are fully expanded and stabilized as described in Section~\ref{sec:vel_curriculum}, we set this value to the final setting of $k=50$, ensuring that the teacher policy adapts well to start with balancing objects instead of directly transporting them. This scheduling strategy balances the need for early gait exploration with the later requirement of stable object transport.
Alongside this schedule, we adjust the probability of sampling standing (zero-command) environments from a relatively higher value during early training to a smaller value in the final stage, ensuring sufficient locomotion diversity while still allowing a fraction of standing cases. Their values in our implementation are shown in Table~\ref{tab:zero_command_schedule}.

\vspace{-0.15cm}
\begin{table*}[h]
    \centering
    \caption{Curriculum schedule for zero velocity commands.}
    \vspace{-0.15cm}
    \begin{tabular}{lc}
    \toprule
        Parameter & Value \\
    \midrule
        Initial number of zero-command steps & $0$ \\
        Final number of zero-command steps & $50$ \\
        Initial probability of sampling standing environments & $0.10$ \\
        Final probability of sampling standing environments & $0.05$ \\
    \bottomrule
    \end{tabular}
    \label{tab:zero_command_schedule}
\end{table*}
\vspace{-0.1cm}

\subsection{Reward Design}
\label{sec:app_reward}

\textbf{Foot Dragging Penalty.}
A key factor leading to early foot landing and dragging is the sim-to-real gap in the material properties of the robot's feet. For Go1 robots, the feet are modeled as rigid bodies in simulation, whereas in reality they are made of rubber, which undergoes significant deformation upon contact. As a result, the deformation of the stance foot causes the swing foot to land earlier in the real world than in simulation, effectively making the ground ``higher'' than the robot perceives in simulation.
Although increasing foot clearance can partially mitigate foot dragging, it mainly addresses obstacle avoidance in rough terrains for blind locomotion policies
and does not fundamentally address the material deformation gap.
To make the robot aware of the virtually ``higher'' ground during simulation, we introduce a foot dragging penalty that penalizes lateral foot motion when the foot is close to the ground, even before actual contact occurs:
\vspace{-0.2cm}
\begin{align*}
r_{\text{drag}}
= 
\sum_{i=1}^{4}
\mathbf{1}_{\left\{
\boldsymbol{p}_{\text{f}_i,z}^{\text{w}} \leq h_{z,\text{th}} \land
\|\boldsymbol{v}_{\text{f}_i,xy}^{\text{w}}\|^2 \geq v_{xy,\text{th}}
\right\}}
\end{align*}
where $f_i$ is the $i$-th foot of the robot, $p^{w}_{i,z}$ is its height above the ground, $\mathbf{v}^{w}_{i,xy}$ is the lateral linear velocity of foot $i$ in the world frame, $h_{z, \text{th}}$ is the height threshold, and $v_{xy,\text{th}}$ is the lateral velocity threshold.

We summarize the reward terms in Table~\ref{tab:reward} and the hyper-parameters for PPO in Table~\ref{tab:ppo}.

\begin{table*}[h]
    \centering
    \caption{
    Reward terms used for training. 
    In the ``object dangerous state'' term, \textbf{danger} condition for object safety is defined as exceeding position or velocity thresholds:
    $|\boldsymbol{p}_{\text{o},x}^{\text{r}}| > x_{\max} \lor 
    |\boldsymbol{p}_{\text{o},y}^{\text{r}}| > y_{\max} \lor 
    |\boldsymbol{p}_{\text{o},z}^{\text{r}}| > z_{\max} \lor 
    \|\boldsymbol{v}_{\text{o},xy}^{\text{r}}\| > v_{xy,\max}$.
    In the ``joint position deviation'' term, \textbf{move} indicates active motion and is defined as 
    $\|\boldsymbol{v}_{\text{cmd}}\|_2 > 0 \lor \|\boldsymbol{v}^{\text{r}}_{xy}\|_2 > v_{\text{th}}$, 
    while \textbf{stand} denotes near-zero commanded and actual base motion: 
    $\|\boldsymbol{v}_{\text{cmd}}\|_2 = 0 \land \|\boldsymbol{v}^{\text{r}}_{xy}\|_2 \le v_{\text{th}}$. 
    }
    \vspace{-0.15cm}
    \renewcommand{\arraystretch}{1.15}
    \setlength{\tabcolsep}{5pt}
    \begin{tabular}{llc}
    \toprule
        Reward & Formulation & Weight \\
    \midrule
        Alive & $\mathbf{1}_{\{\text{episode not terminated}\}}$ & $10.0$ \\
    \midrule
        Robot xy velocity tracking & $\exp\left(-\|\boldsymbol{v}_{xy} - \boldsymbol{v}_{xy}^{\text{cmd}}\|_2 / {\sigma_{vxy}}\right)$ & $1.0$ \\
        Robot yaw velocity tracking & $\exp\left(-|\boldsymbol{w}_{z} - \boldsymbol{w}_{z}^{\text{cmd}}| / {\sigma_{wz}}\right)$ & $0.5$ \\
        Object xy position & $\|\boldsymbol{p}_{\text{o},xy}^{\text{w}} - \boldsymbol{p}_{xy}^{\text{w}}\|_2$ & $-50.0$ \\
        Object yaw angle & $\|\boldsymbol{\theta}_{\text{o},z}^{\text{r}}\|_2$ & $-0.1$ \\
    \midrule
        Foot dragging &
        $\sum_{i=1}^{4}
        \mathbf{1}_{\left\{
        \boldsymbol{p}_{\text{f}_i,z}^{\text{w}} \leq h_{z,\text{th}} \land \|\boldsymbol{v}_{\text{f}_i,xy}^{\text{w}}\|^2 \geq v_{xy,\text{th}}
        \right\}}$ & $-1.0$ \\
        Foot slipping &
        $\sum_{i=1}^{4}
        \|\boldsymbol{v}_{\text{f}_i,xy}^{\text{w}}\| \cdot
        \mathbf{1}_{\{\boldsymbol{F}_{\text{f}_i,z}^{\text{w}} \geq F_{z,\text{th}}\}}$ & $-0.1$ \\
    \midrule
        Adaptive trotting gait &
        $\frac{1}{2}\sum_{(i,j)\in P^{\text{diag}}}
        \gamma_{\text{sym}} \mathbf{1}_{\{c_i=c_j\}}+
        \frac{1}{4}\sum_{(i,j)\in P^{\text{lat}}}\mathbf{1}_{\{c_i\neq c_j\}}$ & $0.5$ \\
    \midrule
        Object z velocity & $\|\boldsymbol{v}_{\text{o},z}^{\text{r}}\|_2$ & $-0.5$ \\
        Object roll angle and velocity & $\|\boldsymbol{\theta}_{\text{o},x}^{\text{r}}\|_2 + \|\boldsymbol{w}_{\text{o},x}^{\text{r}}\|_2$ & $-0.05$ \\
        Object dangerous state & $\mathbf{1}_{\{\text{danger}\}}$ & $-50.0$ \\
        Base height tracking & $|\boldsymbol{p}_{z}^{\text{w}} - h_{\text{target}}|^2$ & $-0.5$ \\
        Base z velocity & $|\boldsymbol{v}_{z}^{\text{w}}|^2$ & $-1.0$ \\
        Base roll-pitch angle & $|\boldsymbol{\theta}_{x}^{\text{w}}|^2 + |\boldsymbol{\theta}_{y}^{\text{w}}|^2$ & $-1.0$ \\
        Base roll-pitch velocity & $|\boldsymbol{w}_{x}^{\text{r}}| + |\boldsymbol{w}_{y}^{\text{r}}|$ & $-0.2$ \\
        Joint position deviation &
        $\|\boldsymbol{q} - \boldsymbol{q}_{\text{default}}\|_2 \cdot 
        \left(
        \mathbf{1}_{\{\text{move}\}} + \alpha_{\text{stance}}\, \mathbf{1}_{\{\text{stand}\}}
        \right)$
        & $-0.5$ \\
        Joint position limits & $\sum_{j=1}^{12} [q^{\min}_j - q_j]_+ + [q_j - q^{\max}_j]_+$ & $-10.0$ \\
        Joint velocity & $\|\dot{\boldsymbol{q}}\|_2$ & $-5\mathrm{e}{-3}$ \\
        Joint acceleration & $\|\ddot{\boldsymbol{q}}\|_2$ & $-5\mathrm{e}{-6}$ \\
        Joint torque & $\|\boldsymbol{\tau}\|_2$ & $-2.5\mathrm{e}{-4}$ \\
        Action rate & $\sum_{j=1}^{12} \|\mathbf{q}_{j, \text{target}} - \mathbf{q}_{j, \text{last target}}\|$ & $-0.75$ \\
        Thigh–calf collision & $\sum_{i=1}^{4} \mathbf{1}_{\{\|\boldsymbol{F}_{t_i}\|_2 > f_{\text{th}}\}} + \mathbf{1}_{\{\|\boldsymbol{F}_{c_i}\|_2 > f_{\text{th}}\}}$ & $-5.0$ \\
    \bottomrule
    \end{tabular}
    \label{tab:reward}
\end{table*}

\begin{table*}[h]
    \centering
    \caption{PPO Hyperparameters.}
    \vspace{-0.15cm}
    \begin{tabular}{lc}
    \toprule
        Parameter & Value \\
    \midrule
        PPO clip range & $0.2$ \\
        GAE discount factor $\lambda$ & $0.95$ \\
        Reward discount factor $\gamma$ & $0.99$ \\
        Learning rate & $1.0\mathrm{e}{-3}$ \\
        Maximum gradient norm & $1.0$ \\
        Entropy coefficient & $0.01$ \\
        Desired KL-divergence & $0.01$ \\
        Number of environment steps per training batch & $24$ \\
        Learning epochs per training batch & $5$ \\
        Number of mini-batches per training batch & $4$\\
        \bottomrule
    \end{tabular}
    \label{tab:ppo}
\end{table*}

\newpage
\subsection{Symmetricity Function}
\begin{figure}[h]
    \centering
    \includegraphics[width=0.48\linewidth]{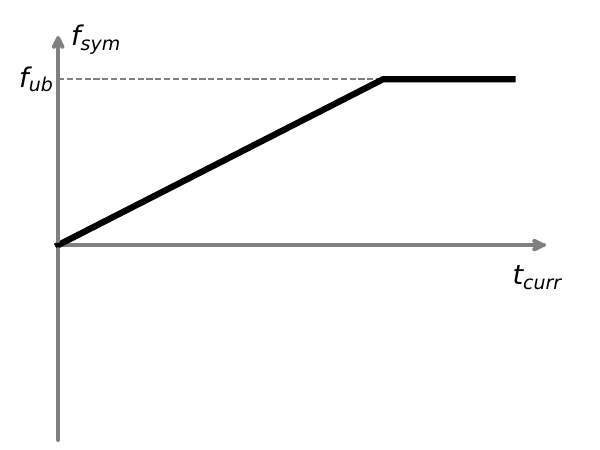}
    \includegraphics[width=0.48\linewidth]{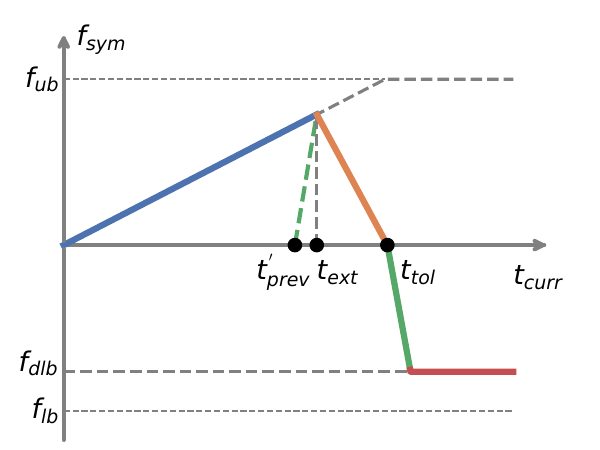}
    \caption{Illustrations of the symmetricity function. \textbf{Left}: when $t_{\text{diff}}\,\leq0$, $f_{\text{sym}}(t)=\alpha_1\,t$ encourages longer air time of the current foot pair. \textbf{Right}: when $t_{\text{diff}}\,>0$, $f_{\text{sym}}$ restrains long air time of the current pair with coordinated sub-functions.}
    \label{fig:symmetry_function}
\end{figure}
Instead of relying on a fixed stepping frequency to compute the frequency tracking reward, we use the previous air time of the alternative foot pair, denoted as $t'_{\text{prev}}$, as the reference for the current foot pair. Based on this reference, we define a symmetricity function $f_{\text{sym}}\left(t_{\text{curr}}\,|\,t_{\text{prev}}, t'_{\text{prev}}\right)$, which evaluates the symmetry factor $\gamma_{\text{sym}}$ from the mean air time $t_{\text{curr}}$ of the current pair, conditioned on both foot pairs’ previous air times $t_{\text{prev}}$ and $t'_{\text{prev}}$, as shown in Figure~\ref{fig:symmetry_function}.

The profile of $f_{\text{sym}}$ is determined by the previous air time difference $t_{\text{diff}} = t_{\text{prev}} - t'_{\text{prev}}$. When $t_{\text{diff}}\,\leq0$, it implies that the current foot pair had a shorter air time in the previous gait cycle. To encourage longer air time in this case, we define $f_{\text{sym}}(t)=\alpha_1\,t$ capped by a global upper bound $f_{\text{ub}}$ (left plot in Figure~\ref{fig:symmetry_function}).

When $t_{\text{diff}}\,>0$, the function is designed to limit excessive air time for the current pair. In this case, $f_{\text{sym}}(t)$ is composed of three linear segments (right plot in Figure~\ref{fig:symmetry_function}), corresponding to the blue, orange, and green lines. The \textbf{first} segment has a positive slope $\alpha_1$, continuing to reward increases in air time up to a tolerance threshold. We define this threshold as $t_{\text{tol}} = (1 + \alpha_{\text{tol}}) t'_{\text{prev}}$, where $\alpha_{\text{tol}}$ is a fixed ratio (0.2 in our implementation). When $t_{\text{curr}} < t_{\text{tol}}$, the symmetry reward remains positive. The positive reward starts to decline at an extended air time $t_{\text{ext}} = t_{\text{tol}} - t_{\text{diff}}$. The \textbf{second} segment, beginning at $t_{\text{ext}}$, has slope $\alpha_2 = f_{\text{sym}}(t_{\text{ext}})/t_{\text{diff}}$, where $f_{\text{sym}}(t_{\text{ext}}) = \alpha_1 t_{\text{ext}}$. The \textbf{third} segment has a negative slope $\alpha_3 = -f_{\text{sym}}(t_{\text{ext}})/(t_{\text{ext}}-t'_{\text{prev}})$ , corresponding to the gradient of the green dotted line. This segment is clamped by a dynamic lower bound $f_{\text{dlb}} = (t_{\text{diff}}/(\alpha_{\text{tol}} t'_{\text{prev}}))\,f_{\text{lb}}$, where $f_{\text{lb}}$ denotes the global lower bound of the reward function.

To discourage naive reduction of stepping frequency, we scale the positive part of the symmetry factor using a task-weighted coefficient: $\gamma_{\text{sym}} = \alpha_{\text{task}} f_{\text{sym}}(t)$, where $\alpha_{\text{task}} \in [0, 1]$ is computed based on task performance. Negative values of $f_{\text{sym}}(t)$ are assigned directly to $\gamma_{\text{sym}}$ without rescaling for preserve strong symmetry constraints.

While $f_{\text{sym}}$ tolerates alternating increases and decreases in air time, another key mechanism for encouraging adaptive stepping is the resetting of $t_{\text{prev}}$ and $t'_{\text{prev}}$ whenever the velocity command changes. Furthermore, the three segments of $f_{\text{sym}}$ act in concert to enforce gait symmetry: for instance, if the reference air time remains constant, a larger $t_{\text{diff}}$ results in a smaller $t_{\text{ext}}$, a lower value of $f_{\text{sym}}(t_{\text{ext}})$, a steeper negative slope $\alpha_3$, and a tighter lower bound $f_{\text{dlb}}$, all contributing to stronger penalties for asymmetric swings.

\newpage
\section{Details for Tactile Policy Distillation}
During distillation, we disable the velocity curriculum and set the zero-command step number $k=50$.
We summarize the tactile policy structure in Table~\ref{tab:student_structure} and the hyper-parameters for distillation in Table~\ref{tab:distill_params}.

\begin{table}[h]
\centering
\begin{minipage}[t]{0.48\textwidth}
    \centering
    \caption{Tactile Policy Structure.}
    \vspace{0.1cm}
    \begin{tabular}{lc}
    \toprule
    \multicolumn{2}{c}{\textbf{Tactile CNN}} \\
    \cmidrule(rl){1-2}
    CNN channels            & $[24, 24, 24]$ \\
    CNN kernel sizes        & $[4, 3, 2]$ \\
    CNN stride              & $[1, 1, 1]$ \\
    CNN pooling layer       & MaxPool \\
    CNN embedding dims      & $64$ \\
    \midrule
    \multicolumn{2}{c}{\textbf{Tactile RNN}} \\
    \cmidrule(rl){1-2}
    RNN type                & GRU \\
    RNN hidden dim          & $512$ \\
    MLP hidden sizes        & $[256, 128, 64, 64]$ \\
    MLP activation          & ELU \\
    \midrule
    \multicolumn{2}{c}{\textbf{Student Backbone}} \\
    \cmidrule(rl){1-2}
    MLP hidden sizes        & $[512, 256, 128]$ \\
    MLP activation          & ELU \\
    \bottomrule
    \end{tabular}
    \label{tab:student_structure}
\end{minipage}
\hfill
\begin{minipage}[t]{0.48\textwidth}
    \centering
    \caption{Distillation Hyperparameters.}
    \vspace{0.1cm}
    \begin{tabular}{lc}
    \toprule
    \multicolumn{2}{c}{\textbf{Environment}} \\
    \cmidrule(rl){1-2}
    Environment number & $405$ \\
    Max eposide steps & $500$ \\
    \midrule
    \multicolumn{2}{c}{\textbf{Training Parameters}} \\
    \cmidrule(rl){1-2}
    Initial BC steps & $400K$ \\
    Incremental DAgger steps & $200K$ \\
    Steps for each batch & $20K$ \\
    Initial BC epochs & $2000$ \\
    Incremental DAgger epochs & $500$ \\
    Learning rate & $5.0\mathrm{e}{-4}$ \\
    \midrule
    \multicolumn{2}{c}{\textbf{Tactile Simulation}} \\
    \cmidrule(rl){1-2}
    Frequency & $40$Hz \\
    Min delay & $0.025s$ \\
    Max delay & $0.05s$ \\
    \bottomrule
    \end{tabular}
    \label{tab:distill_params}
\end{minipage}
\end{table}

\section{Supplementary Experiments in the Real World}

\subsection{System Setup}
\begin{figure}[h]
  \centering
  \includegraphics[width=0.95\linewidth]{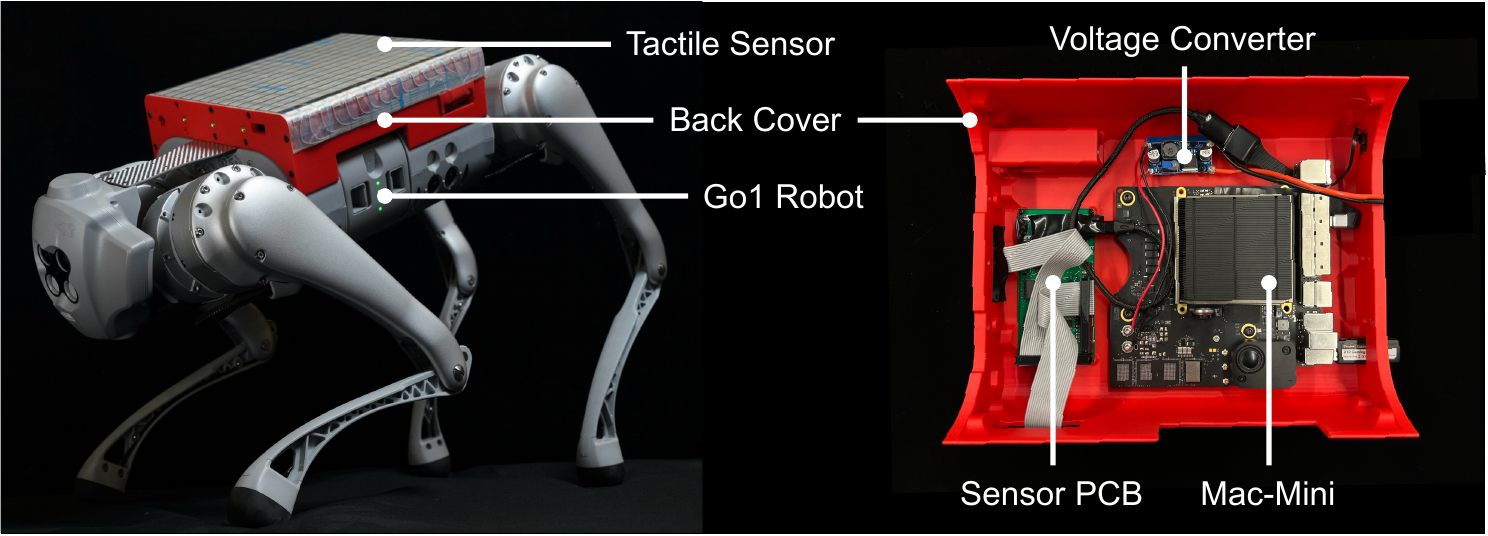}
  \caption{System setup in the real world.}
  \label{fig:system_setup}
\end{figure}

As shown in Figure~\ref{fig:system_setup}, we design and 3D print a custom back cover to mount the tactile sensor on the back of a Unitree Go1 quadrupedal robot. Inside the cover are a PCB (with an onboard Arduino), a Mac Mini computer, and a voltage converter. The PCB reads signals from the tactile sensor. The Mac Mini runs three parallel threads: one for acquiring tactile data via the PCB, one for receiving joystick commands, and one for executing the tactile policy at 50Hz. The resulting motor commands are transmitted to the robot over Ethernet. The voltage converter supplies 12.6V power to the Mac Mini by stepping down the 24V power from the robot.

\newpage
\subsection{Diverse Cylindrical Objects Transport}
\label{sec:app_exp_diverse_obj}
\begin{figure*}[h]
    \centering
    \begin{minipage}{0.5\linewidth}
        \centering
        \includegraphics[width=\linewidth]{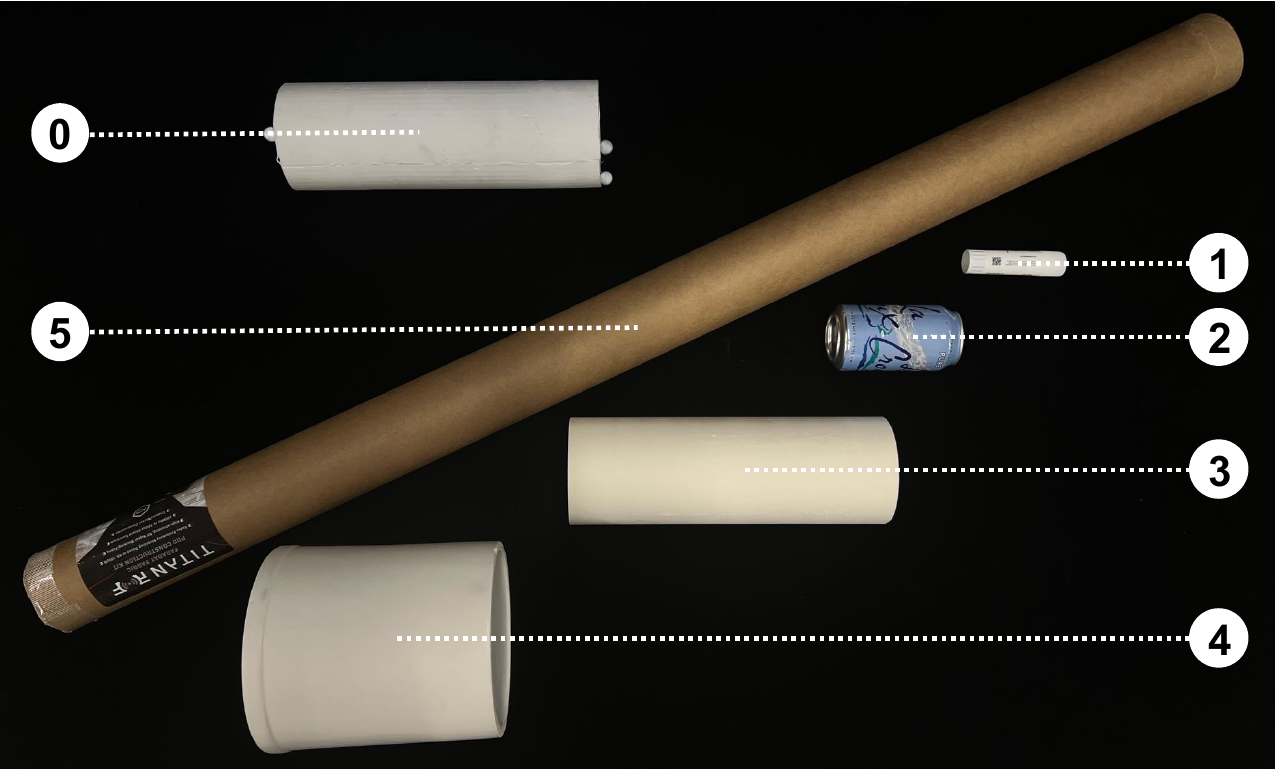}
        \caption{Objects used in our experiments. They have obvious difference in length, diameter, and weight.}
        \label{fig:teaser_objects}
    \end{minipage}%
    \hfill
    \begin{minipage}{0.48\linewidth}
        \centering
        \vspace{-0.2cm}
        \caption{Dimensions and weight of the objects used in our experiments. The success rate is evaluated on a transport distance of $6m$.}
        \label{tab:teaser_objects}
        \renewcommand{\arraystretch}{1.1}
        \setlength{\tabcolsep}{3pt}
        \begin{tabular}{ccccc}
        \toprule
        Object & Length & Diameter & Weight & Success \\
        ID & (m) & (m) & (kg) & Rate \\
        \midrule
        0 & 0.30 & 0.10 & 1.45 & 5/5 \\
        1 & 0.10 & 0.03 & 0.03 & 5/5 \\
        2 & 0.12 & 0.06 & 0.35 & 5/5 \\
        3 & 0.31 & 0.10 & 0.89 & 5/5 \\
        4 & 0.20 & 0.18 & 1.36 & 5/5 \\
        5 & 1.26 & 0.08 & 1.11 & 5/5 \\
        \bottomrule
        \end{tabular}
    \end{minipage}
\end{figure*}

\textbf{Object Properties.}
The objects transported by \nickname in Figure~\ref{fig:teaser} are shown in Figure~\ref{fig:teaser_objects}, with their physical properties listed in Table~\ref{tab:teaser_objects}. These objects vary widely in dimension, mass, and surface friction, making them particularly difficult to balance and transport stably. For example, the glue stick (ID 1) is small and lightweight, making it easy to slip or roll during movement. The drink bottle (ID 2) has a smooth, slippery surface and a varying internal dynamics. The vase (ID 4) is large in diameter and heavy, increasing the risk of sliding or toppling. The poster tube (ID 5) is exceptionally long at $1.26,\mathrm{m}$, nearly six times the width of the robot’s back. The standard cylinder object (ID 0) is attached with some markers to record its state during transport as demonstrated in Figure~\ref{fig:tracking_combined}.

\textbf{Real-World Evaluation.}
We evaluate the transport performance of \nickname on the aforementioned objects in the real world. Each object is randomly placed on the robot’s back within the pose distribution in Table~\ref{tab:randomization}. The robot is commanded to move forward at a velocity of $0.3\mathrm{m/s}$, and transport is considered successful if the object remains stable over a distance of $6\mathrm{m}$. Each test is repeated five times per object. As shown in Table~\ref{tab:teaser_objects}, \nickname achieves a $100\%$ success rate for all objects under the evaluated conditions. However, outside of the evaluation setting, transport may fail if the object is placed in an extremely unfavorable initial pose, for example, when the cylinder’s axis is aligned with the robot’s forward direction.

\subsection{Generalization to Non-Cylindrical Everyday Objects}
\begin{figure}[h]
  \centering
  \includegraphics[width=\linewidth]{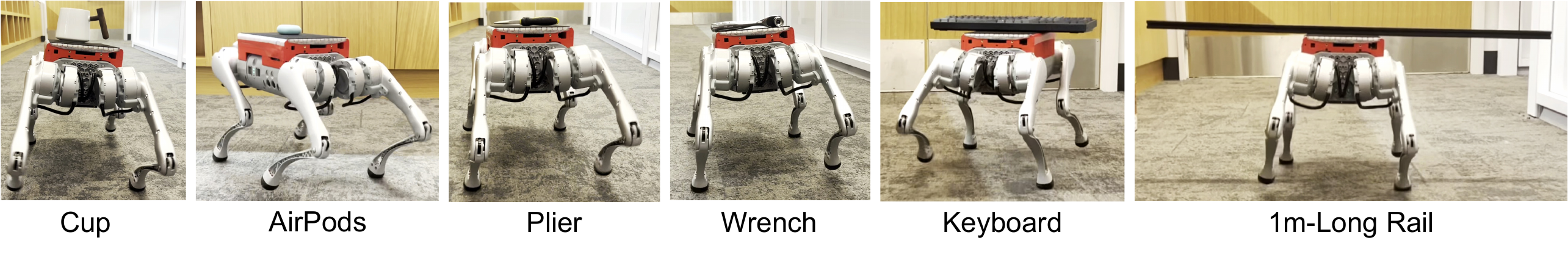}
  \caption{\nickname balances and transports everyday objects such as cups and wrenches.}
  \label{fig:object_generalization}
\end{figure}

Our experiments primarily focus on cylindrical objects, as they represent the most challenging case for transport. 
Nevertheless, we also evaluate \nickname on a variety of non-cylindrical everyday objects, such as cups and wrenches (Figure~\ref{fig:object_generalization}), and find that it can stably balance and transport them. 
This indicates that \nickname does not overfit to the contact patterns of cylindrical objects, likely benefiting from the diverse and unstructured contact patterns generated by the dynamic interactions from cylinders during training.
Notably, when we place a cup slightly off-center on the robot’s back, \nickname is able to actively adjust it toward the center during transport. 
Corresponding videos are available on the~\href{\website}{project website}.

\subsection{Robustness in Uneven Terrains}
\begin{figure}[h]
  \centering
  \includegraphics[width=\linewidth]{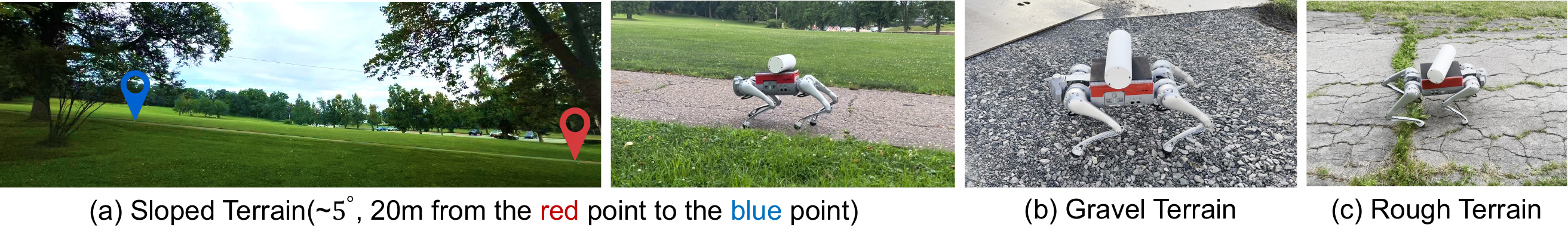}
  \caption{\nickname generalizes to diverse uneven terrains, including slopes, gravel, and rough surfaces.}
  \label{fig:terrain_generalization}
\end{figure}

Although \nickname is trained only on flat terrain in simulation, we find that it generalizes well to diverse environments, including slopes, gravel, and rough surfaces (Fig.~\ref{fig:terrain_generalization}). 
This robustness is likely attributed to the simulated perturbations and domain randomization introduced during training. 
Nevertheless, we believe that incorporating visual perception represents a promising direction to further enhance reliability in more challenging terrains.

\subsection{Long-Distance Transport}
\begin{figure}[h]
  \centering
  \vspace{-0.2cm}
  \includegraphics[width=\linewidth]{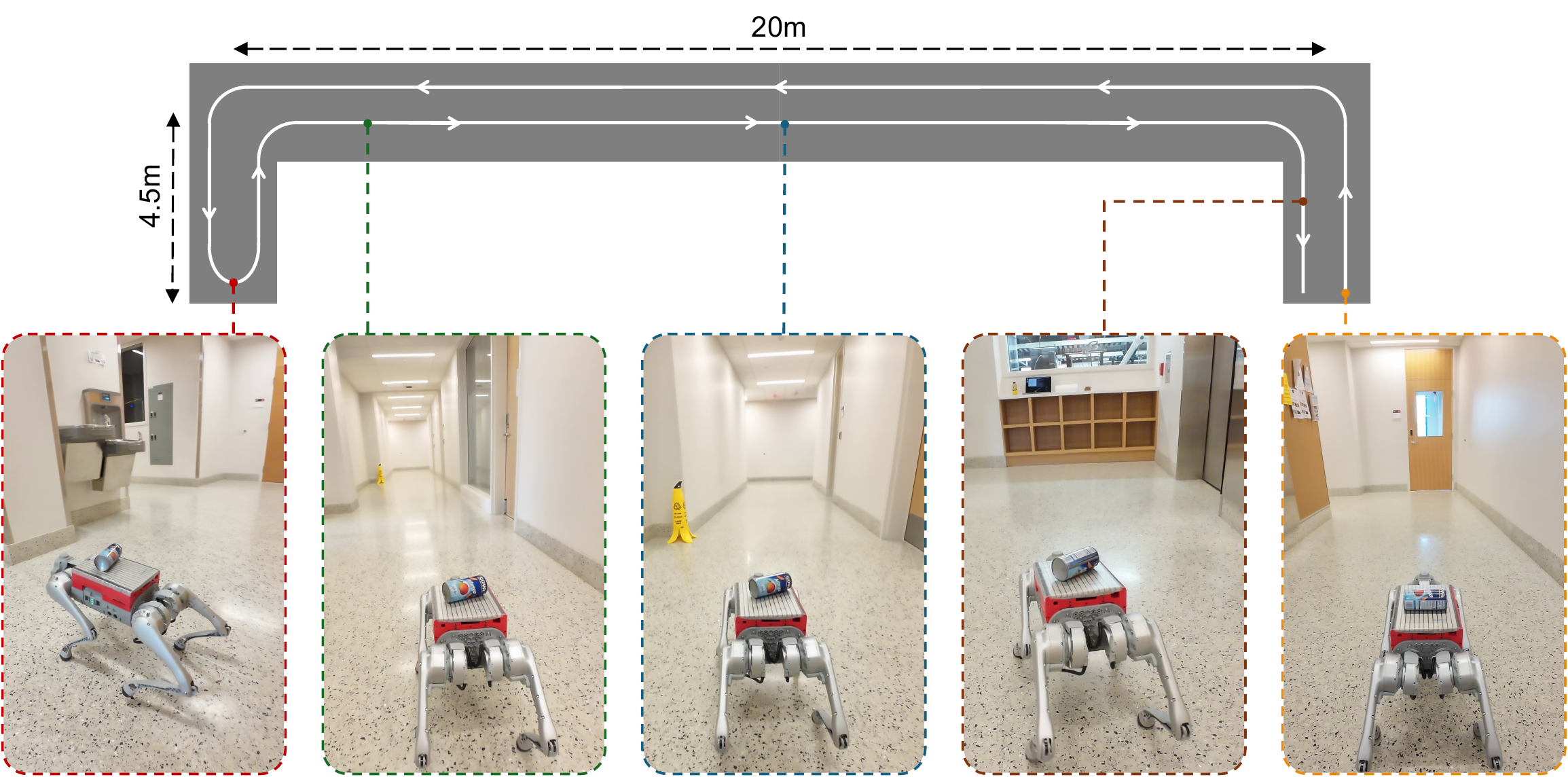}
  \caption{The robot is capable of transporting a slippery drink bottle over long distances while consistently maintaining the object near the center of its back.}
  \label{fig:long_distance}
\end{figure}

We demonstrate that \nickname is capable of transporting objects over long distances, even under varying velocity commands and sharp turns. As shown in Figure~\ref{fig:long_distance}, after carrying the slippery drink bottle for approximately $60\mathrm{m}$, the robot continues to maintain the object near the center of its back. This highlights the robustness of \nickname during real-world deployment.

\subsection{Performance of the Locomotion-Only Policy}
\begin{figure}[th]
  \centering
  \includegraphics[width=\linewidth]{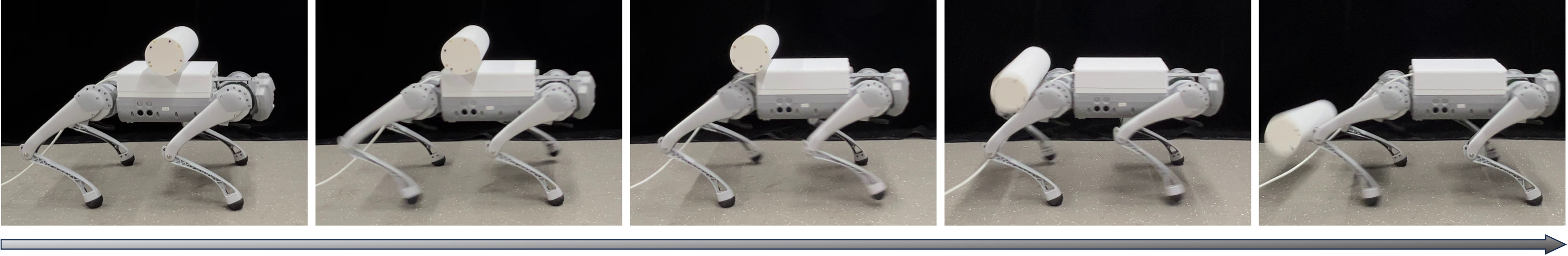}
  \caption{The locomotion policy trained without object interaction totally fails to transport cylindrical objects.}
  \label{fig:baseline_locomotion}
\end{figure}

As shown in Figure~\ref{fig:baseline_locomotion}, the locomotion-only policy trained without interacting with objects cannot even transport the object for $0.5m$.

\end{document}